\documentclass[sigconf, nonacm]{acmart}
\usepackage{graphicx}
\usepackage{balance}
\usepackage{amssymb}
\usepackage{lipsum}
\usepackage{url}
\usepackage{multirow}
\usepackage{hyperref}
\usepackage{pgfplots}
\usepackage{subfiles}
\usepackage{comment}
\usepackage{enumerate}
\usepackage{bbding}
\usepackage{diagbox}
\usepackage{makecell}
\newcommand\vldbdoi{XX.XX/XXX.XX}
\newcommand\vldbpages{XXX-XXX}
\newcommand\vldbvolume{0}
\newcommand\vldbissue{0}
\newcommand\vldbyear{2024}

\newcommand\vldbavailabilityurl{URL_TO_YOUR_ARTIFACTS}
\newcommand\vldbpagestyle{plain}

\begin{document}
\title{A Comprehensive Survey of Dynamic Graph Neural Networks: Models, Frameworks, Benchmarks, Experiments and Challenges 
}

\author{ZhengZhao Feng$^1$, Rui Wang$^{1,2,*}$, TianXing Wang$^1$, Mingli Song$^{1,3}$, Sai Wu$^{2,1}$, Shuibing He$^1$}
\affiliation{
    \institution{$^1$ Zhejiang University, Hangzhou, China\\
    $^2$ Hangzhou High-Tech Zone (Binjiang) Institute of Blockchain and Data Security, Hangzhou, China\\
    $^3$ Shanghai Institute for Advanced Study, Zhejiang University, Shanghai, China
    }
}
\email{{fengzhengzhao, rwang21, tianxingwang, brooksong, wusai, heshuibing}@zju.edu.cn}

\begin{abstract}
Dynamic Graph Neural Networks (GNNs) combine temporal information with GNNs to capture structural, temporal, and contextual relationships in dynamic graphs simultaneously, 
leading to enhanced performance in various applications. 
As the demand for dynamic GNNs continues to grow, numerous models and frameworks have emerged to cater to different application needs. 
There is a pressing need for a comprehensive survey that evaluates the performance, strengths, and limitations of various approaches in this domain.  
This paper aims to fill this gap by offering a thorough comparative analysis and experimental evaluation of dynamic GNNs. 
It covers 81 dynamic GNN models with a novel taxonomy, 12 dynamic GNN training frameworks, and commonly used benchmarks. 
We also conduct experimental results from testing representative nine dynamic GNN models and three frameworks on six standard graph datasets. 
Evaluation metrics focus on convergence accuracy, training efficiency, and GPU memory usage, enabling a thorough comparison of performance across various models and frameworks.  
From the analysis and evaluation results,
we identify key challenges and offer principles for future research to enhance the design of models and frameworks in the dynamic GNNs field.
\end{abstract}

\maketitle

\pagestyle{\vldbpagestyle}
\begingroup\small\noindent\raggedright\textbf{PVLDB Reference Format:}\\
ZhengZhao Feng, Rui Wang, TianXing Wang, Mingli Song, Sai Wu, Shuibing He. 
A Comprehensive Survey of Dynamic Graph Neural Networks: Models, Frameworks, Benchmarks, Experiments and Challenges. 
PVLDB, \vldbvolume(\vldbissue): \vldbpages, \vldbyear. 
\href{https://doi.org/\vldbdoi}{doi:\vldbdoi}
\endgroup
\begingroup
\renewcommand\thefootnote{}\footnote{\noindent
*Corresponding author. \\
This work is licensed under the Creative Commons BY-NC-ND 4.0 International License. Visit \url{https://creativecommons.org/licenses/by-nc-nd/4.0/} to view a copy of this license. For any use beyond those covered by this license, obtain permission by emailing \href{mailto:info@vldb.org}{info@vldb.org}. Copyright is held by the owner/author(s). Publication rights licensed to the VLDB Endowment. \\
\raggedright Proceedings of the VLDB Endowment, Vol. \vldbvolume, No. \vldbissue\ %
ISSN 2150-8097. \\
\href{https://doi.org/\vldbdoi}{doi:\vldbdoi} \\
}\addtocounter{footnote}{-1}\endgroup

\ifdefempty{\vldbavailabilityurl}{}{
\vspace{.3cm}
\begingroup\small\noindent\raggedright\textbf{PVLDB Artifact Availability:}\\
The source code, data, and/or other artifacts have been made available at \url{https://github.com/fengwudi/DGNN_model_and_data}.
\endgroup
}

\section{Introduction}
\label{sec:intro}

Graphs are essential for representing, analyzing, interpreting, and predicting real-world phenomena~\cite{future_graph}. 
Graph neural networks (GNNs), such as GCN~\cite{GCN}, GraphSAGE~\cite{GraphSAGE} and GAT~\cite{GAT}, combine traditional graph computation with deep learning techniques, 
achieving success in tasks like link prediction~\cite{link_prediction}, node classification~\cite{gnn_survey1}, and attribute prediction~\cite{attribute_prediction}.
To enhance the extraction of real-world dynamic graph insights, 
dynamic GNNs (DGNNs) like EvolveGCN~\cite{EvolveGCN}, T-GCN~\cite{T-GCN}, JODIE~\cite{JODIE}, and TGN~\cite{TGN} 
integrate temporal information with GNNs to capture structural, temporal, and contextual relationships within dynamic graphs~\cite{temporal_graph_study}. 
These dynamic GNN models surpass static GNN models in many tasks 
with potential applications in social network analysis~\cite{social_network_analysis}, time series prediction~\cite{JODIE}, and traffic flow forecasting~\cite{T-GCN}.

Dynamic GNNs or temporal GNNs have gained significant attention in the literature~\cite{Kazemi_survey}. 
A plethora of DGNN models and frameworks have been created to 
address various applications~\cite{STGCN, DEGC, Netwalk, RT-GCN, TFE-GNN, TNDCN}, 
enhance inference accuracy~\cite{WD-GCN, EvolveGCN, TeMP, PI-GNN}, 
and improve training efficiency~\cite{TNA, HTGN, SEIGN, SSGNN, SpikeNet}.
While several surveys on dynamic GNN models exist, they primarily focus on algorithms that fit the encoder-decoder architecture~\cite{Encoder-Decoder_survey}.  For instance, 
\cite{Kazemi_survey} review representation learning techniques for dynamic graphs, 
\cite{Skarding_survey} explores the application of DGNN models in dynamic graph analysis, 
and \cite{zhu_survey} proposes a three-stage recursive temporal learning framework based on dynamic graph evolution theory.  
Furthermore, \cite{jin_survey} exclusively concentrates on spatio-temporal graphs, 
while \cite{longa_survey} presents a unique classification approach for DGNN models but with a limited scope.  
Although these surveys offer valuable insights, 
they are constrained to a narrow subset of the DGNN development landscape, 
exhibiting certain limitations:

\noindent\textbf{L1: Outdated coverage of research.} 
The surveys by \cite{Kazemi_survey, Skarding_survey, zhu_survey} are somewhat dated, missing out on the numerous new DGNN models that have emerged after their publication. 
On the other hand, while \cite{jin_survey} and \cite{longa_survey} provide more current perspectives, the former concentrates solely on spatio-temporal graphs, and the latter examines only a restricted range of DGNN models.

\noindent\textbf{L2: Absence of discussion on DGNN frameworks.} 
Apart from a brief mention in \cite{longa_survey} regarding TGL, none of the surveyed works discuss DGNN frameworks. 
Yet, these frameworks are essential for integrating models, optimizing training, and improving performance and scalability.
They serve as a centralized platform for crafting various DGNN architectures, integrating efficient data processing, parallel computation, and tailored optimization algorithms for dynamic graph data. 
Moreover, DGNN frameworks facilitate advanced functionalities such as distributed training, crucial for handling large-scale datasets and real-world applications.

\noindent\textbf{L3: Oversight of evaluation benchmarks.}  
These surveyed works lack a detailed overview of evaluation benchmarks. 
While such benchmarks may be known to experienced researchers, newcomers may benefit from a comprehensive explanation.  
It is essential to introduce evaluation benchmarks thoroughly, including definitions and usage methods, to assist readers in understanding these criteria.

\noindent\textbf{L4: Lack of experimental comparisons.} 
Existing surveyed works serve as foundational overview of DGNN models but lacks experimental comparisons among these models. 
Such comparisons are vital for grasping performance discrepancies across various models.
Challenges arise in unequivocally ranking DGNN models due to differences in datasets and evaluation metrics. 
Some models excel on specific datasets but may falter on others, lacking scalability. 
Moreover, variations in experimental settings can yield different results. 
Therefore, a standardized experimental setup with comprehensive and fair comparisons is necessary to tackle these issues.

\noindent\textbf{L5: Meeting emerging application demands and new challenges.} 
As technology advances, DGNN models have the potential to revolutionize various emerging fields, albeit not yet fully leveraged. 
Previous research has addressed existing challenges to some extent; 
however, the emergence of new application demands poses fresh hurdles for Dynamic GNNs to navigate.

In response to the limitations mentioned above, 
we target to offer a comprehensive and up-to-date overview of dynamic GNNs, encompassing recent research advancements. 
We introduce new classification methods to adapt to the evolving landscape of dynamic GNN models and explore existing frameworks and evaluation benchmarks. 
Additionally, we conduct thorough experimental comparisons of prominent DGNN models and frameworks, and examine emerging application demands and challenges in the field of dynamic GNNs. 
Our main contributions are outlined as follows:

\noindent\textbf{C1: Comprehensive survey and novel taxonomy of DGNN models.}  
Addressing the limitation mentioned in \textbf{L1}, 
we conducted an extensive survey of 81 recent DGNN models and introduced a novel classification approach \textbf{(\S\ref{sec:models})}. 
This taxonomy provides unique insights and perspectives in the current research field. 
By categorizing DGNN models according to their structures, features, and dynamic modeling methods, we facilitate a better understanding and comparison of the strengths and weaknesses of each model.

\noindent\textbf{C2: Overview of existing DGNN frameworks.}  
For limitation \textbf{L2}, 
we provide a detailed overview of current 12 DGNN frameworks, exploring their features and improvements in model optimization \textbf{(\S\ref{sec:frameworks})}. 
Our emphasis on the frameworks' flexibility, scalability, and performance underscores their essential characteristics.

\noindent\textbf{C3: Introduction of evaluation benchmarks for DGNN.}  
To address \textbf{L3}, we present a diverse set of commonly used evaluation graph datasets and metrics for the models discussed \textbf{(\S\ref{sec:benchmarks})}. 
This extensive coverage aims to facilitate comprehensive evaluations and enhance reproducibility in various experiments.

\noindent\textbf{C4: Experimental comparison of selected works.}  
To address \textbf{L4}, we conduct a comprehensive comparison of various DGNN models and frameworks under consistent experimental settings, datasets, and metrics \textbf{(\S\ref{sec:experiments})}. 
We evaluate training accuracy, efficiency, and memory usage of these models and frameworks. 
We also assess multi-GPU scalability within the frameworks.  

\noindent\textbf{C5: Analysis of challenges in DGNN.}  
To address \textbf{L5}, we analyze the new challenges encountered by DGNN models and frameworks and suggest potential research directions for practitioners \textbf{(\S\ref{sec:challenges})}.

\section{Background}
\label{sec:background}


\subsection{Applications of Dynamic Graph Learning}
\label{subsec:applications}

\subsubsection{Dynamic Graph Scenarios}
Real-world graphs exhibit dynamic features and find applications in various domains, including:

\noindent\textbf{Temporal Interaction Graphs in Social Networks:} 
Temporal interaction graphs capture the evolving relationships and interactions between social network users over time, offering insights into social dynamics and network evolution.  

\noindent\textbf{Real-Time Transaction Graphs in E-Commerce:} 
Transaction graphs model the flow of transactions and interactions between users and products in real-time, facilitating fraud detection, recommendation systems, and personalized marketing strategies.  

\noindent\textbf{Spatio-Temporal Graphs (STG):} 
These graphs integrate spatial and temporal dimensions, with nodes representing spatial locations and edges denoting spatial relationships. 
They enable the analysis of movement patterns, traffic flow, and environmental changes, aiding urban planning, transportation management, and environmental monitoring.  
Temporal information for nodes and edges can enhance the understanding of time-dependent spatial relationships. 

\noindent\textbf{Temporal Knowledge Graphs (TKG):}  
Knowledge graphs depict structured information, with entities as nodes and relationships as edges.  
Adding a temporal dimension in the triplets, TKGs allow to track the changes in entities and relationships over time, essential for applications like online social networks and trend analysis.  

\noindent\textbf{Temporal Citation Graphs (TCG):} 
TCGs track the evolution of citations and references in scholarly publications over time, enabling analyses of research trends, influence dynamics, and knowledge dissemination patterns within academic fields.

\subsubsection{Learning Tasks in Dynamic Graphs}
Dynamic graph learning can aid in various tasks in above application domains, including:  

\noindent\textbf{Link Prediction:} 
Involves predicting the likelihood of connections between two nodes in a network that do not yet have edges based on existing network nodes and structure. 
In dynamic graph learning, link prediction focuses on forecasting the probability of edges appearing between nodes at a specific time.  

\noindent\textbf{Node Classification:} 
Entails assigning labels to nodes with unknown labels in a graph by utilizing the connections between nodes and a limited number of labeled nodes.  In dynamic graph learning, this task aims to predict the labels of nodes at a given time.  

\noindent\textbf{Other Tasks:} 
Apart from the above two common tasks, 
there are additional specialized tasks, such as 
temporal node embedding which learns low-dimensional representations of nodes that capture the temporal dynamics and structural changes in the graph, 
temporal graph embedding which learns representations of the entire graph at different time steps, 
and event prediction which predicts future events or occurrences in a dynamic graph




\begin{figure*}
	\centering 
	\includegraphics[width=1\textwidth, angle=0]{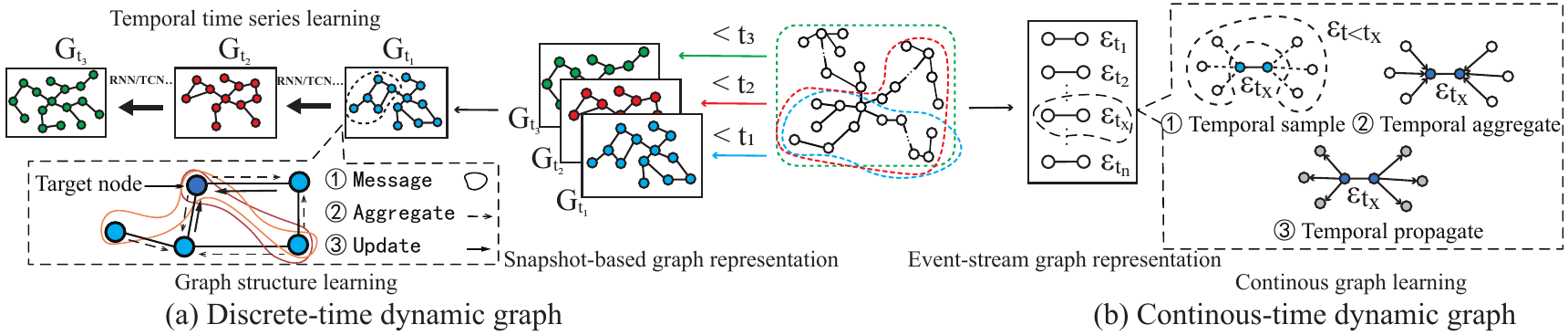}	
    \vspace{-20pt}
	\caption{A toy example of dynamic graph representation and learning.}
	\label{fig:DGNN_train_photo}%
    \vspace{-15pt}
\end{figure*}

\subsection{Representation of Dynamic Graphs}
\label{subsec:representation}

\subsubsection{Graph Structure Representation}
We now discuss the graph structure representation from static graphs to dynamic graphs.

\noindent\textbf{Static Graph Representation:}  
Static graphs are typically represented by nodes, edges, and features: $G=(V,E,X)$, where $V$ is the node set, $E$ is the edge set, and $X$ represents the feature embeddings of the graph.  
An adjacency matrix $A \in \mathbb{R}^{|V| \times |V|}$ is commonly used to depict a static graph. In the matrix, a value of 1 at $A[i][j]$ indicates an edge from node $i$ to node $j$.  
An edge index $Edge_{index}\in \mathbb{R}^{2 \times |E|}$ is also utilized for graph data, with each column representing an edge in the graph. For example, nodes $Edge_{index}[0][i]$ and $Edge_{index}[1][i]$ define the $i$-th edge.

\noindent\textbf{Dynamic Graph Representation:} 
Dynamic graphs add a temporal dimension to static graphs. 
At time $t$, the graph is represented as $G_t=(V_t, E_t, X_t)$, with $V_t$, $E_t$, and $X_t$ representing nodes, edges, and features at that time. 
Two common approaches to dynamic graph representation are {\em discrete-time dynamic graphs (DTDG)} and {\em continuous-time dynamic graphs (CTDG)}. 
The difference between DTDG and CTDG is illustrated in Figure \ref{fig:DGNN_train_photo}.  
In DTDG, the timestamp $T_n=[t_1:t_n]$ is divided into $n$ time intervals, and the dynamic graph is represented as a sequence of graph snapshots within $T_n$ denoted as $G_T=({G_{t_1}, G_{t_2},..., G_{t_n}})$. 
Each $G_{t_i}$ captures the graph structure up to time $t_i$. 
On the other hand, in CTDG, graph information is treated as an event stream $G_T=(\varepsilon_{t_1}, \varepsilon_{t_2}, ..., \varepsilon_{t_n})$. 
An edge created at time $t$ from node $i$ to node $j$ is represented as $\varepsilon_t=(i,j,t)$. 
CTDG maintains a single graph structure at any given time $t$ by incorporating all event stream data to form $G_t$.



\subsubsection{Temporal Time Series Representation}  
In the context of dynamic graphs, temporal time series data can be represented using explicit or implicit time methods.  

\noindent\textbf{Explicit Time Representation:}  
Explicit time involves including time as a distinct feature in the model for computation.  This approach incorporates time series data directly into the model as an input feature, influencing the model's calculations and enabling it to utilize time information for tasks like prediction or learning.  

\noindent\textbf{Implicit Time Representation:}  
On the other hand, implicit time implies that the model comprehends the temporal progression within its structure without explicitly treating time as an input feature.  Instead of requiring specific time values, the model learns temporal relationships through the sequential arrangement of data.  This allows the model to capture temporal evolution from the data without emphasizing timestamps or explicit time features.


\subsection{Learning of Dynamic Graphs}
\label{subsec:learning}

\subsubsection{Graph Structure Learning}
Diverse methods have been developed to handle various types of graph structure data, ranging from neural networks and attention mechanisms to random walks: 

\noindent\textbf{Graph Neural Networks (GNNs):}  
GNNs are neural network models tailored for processing graph data, 
allowing for the effective capture of intricate relationships within graph structures and enabling node and graph learning and inference.  
A fundamental aspect of GNN design is the utilization of {\em message passing mechanisms}, 
which typically involves three key steps: {\em message}, {\em aggregate}, and {\em update}.
Specifically, each vertex first collects feature embedding messages from its neighboring vertices, then aggregates the collected feature messages using an aggregation function, and finally updates the vertex's feature embedding information using neural network model parameters.


\noindent\textbf{Structural Attention:}  
Attention-based graph learning utilizes attention mechanisms in GNNs to prioritize information flow between nodes in a graph. 
This allows the model to focus on key nodes or edges dynamically, adjusting their importance during message passing and feature aggregation.  

\noindent\textbf{Random Walk:}  
Random walk-based graph learning employs random walk algorithms on graphs to capture structural relationships between nodes. 
In this method, walkers move between nodes based on predefined rules (e.g., edge probabilities) to explore the graph. 
By conducting multiple random walks and analyzing visited node sequences, 
we can generate node embeddings that encode the local graph structure and connectivity patterns.

\subsubsection{Temporal Time Series Learning} 
Next, we introduce the learning method for temporal time series information:

\noindent\textbf{Recurrent Neural Networks (RNN):} 
RNN is a classic neural network architecture for processing sequential data like time series and texts. 
It excels at capturing sequence context and adapting to different sequences lengths. Advanced RNN variations like long short-term memory (LSTM)~\cite{LSTM} and gated recurrent unit (GRU)~\cite{GRU} address challenges like gradient vanishing and exploding. 

\noindent\textbf{Temporal Point Process (TPP):} 
TPP~\cite{TPP} is a statistical model used to analyze event patterns over time, 
focusing on event occurrence moments rather than event count or intensity~\cite{tpp1,tpp2}. 
The conditional intensity function $\lambda(t)$ describes the intensity of future events based on historical event information $H_t$ before time $t$. 

\noindent\textbf{Temporal Convolutional Network (TCN):} 
TCN~\cite{TCN} is a deep learning model for time series data modeling. 
Unlike RNNs, TCN employs a convolutional neural network structure to capture dependencies in time series, effectively capturing local dependencies within sequential data.

\noindent\textbf{Temporal Attention:} 
Temporal attention is utilized for processing temporal data, enabling models to assign varying importance to information at different time steps when handling sequential data. 

\noindent\textbf{Time Encoding:} 
Time encoding involves representing temporal information as features for model training. 
One common approach is using parameterized Fourier features\cite{time_encoding}. 


\subsection{Dynamic Graph Neural Networks} 
\label{sebsec:DGNN}  

\subsubsection{DGNN Models on DTDG and CTDG} 
In \S\ref{subsec:representation}, we explored how dynamic graphs can be represented using either DTDG or CTDG, leading to distinct DGNN models. 
The training differences for DGNN models in DTDG and CTDG are depicted in Figure \ref{fig:DGNN_train_photo}. 
While DTDG and CTDG vary in terms of snapshots and event streams, the key contrast lies in the granularity of time representation and learning: coarse-grained time in DTDG and fine-grained time in CTDG.  
In DTDG, a fixed time interval like a month or a year is often selected to partition the graph into multiple snapshots when edges are added over time. 
Conversely, CTDG spans an infinite time range, capturing all event streams in a single snapshot graph with precise time ordering and sequential event processing.









\section{Taxonomy of Dynamic GNN Models}
\label{sec:models}

\begin{figure}[t]
	\centering 
	\includegraphics[width=0.5\textwidth, angle=0]{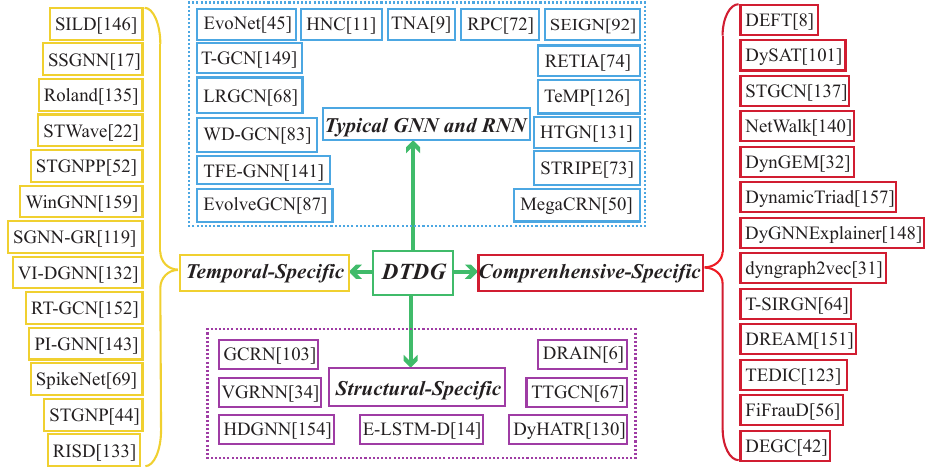}	
    \vspace{-15pt}
	\caption{The taxonomy for DTDG in our survey.}
	\vspace{-15pt}
	\label{fig:DTDG_photo}
\end{figure}

This section offers a detailed overview of the most recent dynamic GNN models, 
covering 48 DTDG models and 33 CTDG models. 
We introduce a new classification method that categorizes these Dynamic GNN models according to their structural features, use of methods, and dynamic modeling techniques. 

\subsection{Discrete-Time Dynamic Graph Models} 
\label{subsec:DTDG}  

DTDG models are designed for analyzing discrete-time dynamic graphs. 
These models can be classified into four groups based on their methods for handling graph structure and temporal data, as illustrated in Figure~\ref{fig:DTDG_photo}. 
\textbf{Typical GNN and RNN models} involve using GNNs for processing graph structure data and RNNs for temporal time series data. 
Other DTDG models fall into the categories of \textbf{structural-specific}, \textbf{temporal-specific}, and \textbf{comprehensive-specific models}, 
depending on their reliance on non-GNN methods for graph structure processing, non-RNN methods for temporal data processing, or neither GNN nor RNN, respectively.

\subsubsection{Typical GNN and RNN Models} 
The conventional approach of utilizing GNN for processing graph structures and RNN for handling temporal information is considered the most intuitive and prevalent method in DGNN modeling.

\noindent\textbf{Enhancing Accuracy.} 
WD-GCN~\cite{WD-GCN} was among the first algorithms to combine GNN and RNN for dynamic graph processing, merging GCN~\cite{GCN} and LSTM~\cite{LSTM} models.  
EvolveGCN~\cite{EvolveGCN} updates the weight matrix of GCN models across different time frames using RNN.  
TeMP~\cite{TeMP} incorporates temporal message passing for improved predictions, blending the structural aspects of Relational GCN and GRU~\cite{GRU}.  
EvoNet~\cite{EvoNet} integrates graph-level propagation with GNN and RNN components to comprehensively capture temporal graph information.  
RETIA~\cite{RETIA} considers adjacent relationships in addition to aggregating neighboring entities, leading to a more holistic learning approach.  
Incorporating GNN, GRU, and novel units RCU and PCU, RPC~\cite{RPC} explores relational correlations and periodic patterns, enhancing the model's expressiveness.

\noindent\textbf{Improving Training Efficiency and Scalability.}  
TNA~\cite{TNA} enhances training stability beyond traditional GCN and GRU structures.  
HTGN~\cite{HTGN} maps the temporal graph onto hyperbolic space and introduces specific modules to constrain the model, ensuring stability and generalization of embeddings.  
SEIGN~\cite{SEIGN} utilizes a three-part structure involving GCN-like message passing, GRU parameter adjustments over time, and a self-attention mechanism inspired by transformers~\cite{transformer} for learning the final representation.  
This strategy enhances scalability and efficiency, enabling effective training on large-scale graphs.

\noindent\textbf{Specialized Applications.}  
Several typical models are tailored for specific application or domains.  
LRGCN~\cite{LRGCN} incorporates LSTM~\cite{LSTM} based on Relational R-GCN~\cite{R-GCN} for path fault prediction.  
T-GCN \cite{T-GCN} is designed for traffic prediction, combining GCN and GRU.  
MegaCRN \cite{MegaCRN} introduces the GCRN unit for processing spatio-temporal graphs, utilizing a hybrid architecture of GCN and GRU.  
TFE-GNN~\cite{TFE-GNN} processes graph structure using multi-layer GNNs.  
HNC~\cite{HNC} manages satellite communication status within a satellite network, aiding in the development of a global satellite joint program.  
STRIPE~\cite{STRIPE} is the first to incorporate memory networks for anomaly detection in spatio-temporal graphs.

\subsubsection{Structural-Specific Models} 
This category of DTDG models utilizes non-GNN methods for processing graph structural data and still uses RNN for temporal time series data processing.   

\noindent\textbf{Enhancing Accuracy.}  
GCRN~\cite{GCRN} combines GCN and RNN to predict dynamic data using both spatial and temporal information. 
VGRNN~\cite{VGRNN} builds upon GCRN by transitioning to GRNN and introducing the dynamic graph auto-encoder model.  
E-LSTM-D~\cite{E-LSTM-D} is the first to utilize LSTM and an encoder-decoder architecture for link prediction in dynamic networks.  
TTGCN~\cite{TTGCN} introduces a dynamic graph representation learning method based on k-truss decomposition, effectively capturing multi-scale topological structure information in graphs.

\noindent\textbf{Improving  Training Efficiency.} 
DRAIN~\cite{DRAIN} has constructed a recurrent graph generation scenario to represent a dynamic GNN that learns across different time points. 
It captures the time drift of model parameters and data distributions, and can predict future models in the absence of future data.

\noindent\textbf{Specialized Applications.}  
Some structural-specific models are tailored for managing heterogeneous graphs. 
HDGNN~\cite{HDGNN} employs multi-head attention and random walk techniques for processing structural information,  
whereas DyHATR~\cite{DyHATR} integrates a hierarchical attention mechanism to learn heterogeneous information.

\subsubsection{Temporal-Specific Models}
This category of DTDG models employs GNN for processing graph structural data and non-RNN methods for temporal time series data processing.  

\noindent\textbf{Enhancing Accuracy.} 
SGNN-GR~\cite{SGNN-GR} utilizes generative adversarial networks (GAN) to generate synthetic time information, addressing catastrophic forgetting issues without additional data storage. Paired with GraphSAGE~\cite{GraphSAGE}, it establishes a framework for DGNN.  
PI-GNN~\cite{PI-GNN} employs parameter isolation for dynamic graphs to capture emerging patterns without compromising older ones.  
STGNP~\cite{STGNP} introduces the spatio-temporal graph neural process, incorporating neural processes~\cite{NP} for modeling spatio-temporal graphs.  
SILD~\cite{SILD} firstly explores distribution shifts in the spectral domain of dynamic graphs. 

\noindent\textbf{Improving Training Efficiency and Scalability.} 
ROLAND~\cite{ROLAND} introduces embedding update modules to handle adjacent time snapshots, enabling the extension of static graphs to dynamic graphs.  
WinGNN~\cite{WinGNN} replaces time encoders with random sliding windows and adaptive gradients, effectively reducing the number of parameters.  
SSGNN~\cite{SSGNN} utilizes echo state networks (ESN)~\cite{ESN} to enhance scalability for large networks.  
SpikeNet~\cite{SpikeNet} replaces RNN functionality with spiking neural networks (SNN)~\cite{SNN}, reducing model and computational complexity through the leaky integrate-and-fire (LIF) model of SNN.

\noindent\textbf{Specialized Applications.} 
RT-GCN~\cite{RT-GCN} predicts stocks using relationship and time convolution, employing GCN for relationship convolution and TCN for time convolution. 
RISD~\cite{RISD} is designed for heterogeneous graphs, utilizing heterogeneous GCN to learn graph representations and construct sampling graphs. 
STWave~\cite{STWave} is a framework for traffic prediction, incorporating GAT for spatial dimensions and a transformer for temporal dimensions to identify spatio-temporal correlations. 
STGNPP~\cite{STGNPP} combines GCN and transformer for predicting traffic congestion time using neural process priors (NPP). 
VI-DGNN~\cite{VI-DGNN} presents a model tailored for trajectory prediction tasks in the intelligent transportation domain.

\subsubsection{Comprehensive-Specific Models}
The final category comprises models that do not fall into the previous classifications, 
which neither use GNN nor RNN for dynamic graph processing.

\noindent\textbf{Enhancing Accuracy.} 
Dyngraph2vec~\cite{dyngraph2vec} is an early Dynamic GNN model capable of capturing temporal dynamics. 
DynGEM~\cite{DynGEM} initializes snapshot embeddings from previous time steps before gradient training, avoiding learning from scratch. 
DySAT~\cite{DySAT} incorporates an attention mechanism to learn structural and temporal information efficiently compared to RNN-based solutions.  

\noindent\textbf{Improving Scalability.}  
T-SIRGN~\cite{T-SIRGN}, inspired by SIR-GN~\cite{SIRGN}, extends the capabilities of existing approaches to handle efficiency and scalability issues in dynamic graphs.

\noindent\textbf{Path-based Methods.} 
DynamicTriad~\cite{DynamicTriad} utilizes triples for link prediction to capture network dynamics. 
Netwalk~\cite{Netwalk} reconstructs node representations and groups embeddings along walking paths using a deep auto-encoder model.  

\noindent\textbf{Capturing Global Information.} 
DEFT~\cite{DEFT} employs learning spectral wavelets~\cite{wavelet} to capture global features, effectively learning dynamic evolution graph representations. 
DREAM~\cite{DREAM} captures long-term evolution through a temporal self-attention network. 
EAGLE~\cite{EAGLE} utilizes the {\em modeling-inferring-discriminating-generalizing} paradigm to enhance extrapolation capabilities in the future.

\noindent\textbf{Specialized Applications.} 
STGCN~\cite{STGCN} focuses on spatio-temporal correlations for long-term traffic prediction tasks.  
TEDIC~\cite{TEDIC} models information flow using enhanced graph convolution and extracts fine-grained patterns over time for dynamic social interaction pattern extraction.  
DEGC~\cite{DEGC} addresses recommendation system challenges with an isolation-based approach to handle obsolescence.  
FiFrauD~\cite{FiFrauD} is an unsupervised and scalable method for detecting suspicious traders and behavioral patterns efficiently.  
DyGNNExplainer~\cite{DyGNNExplainer} introduces a causality-inspired generative model based on structural causal models to explore the philosophy of DyGNN prediction by identifying trivial, static, and dynamic causal relationships.

\begin{figure}[b]
	\centering 
	\includegraphics[width=0.5\textwidth, angle=0]{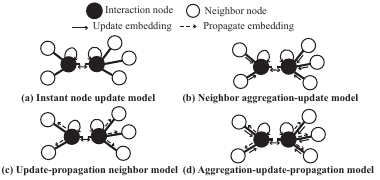}	
	\vspace{-20pt}
	\caption{Node update methods based on event streams.}
	\label{CTDG_update_photo}%
\end{figure}

\begin{figure}[t]
	\centering 
	\includegraphics[width=0.5\textwidth, angle=0]{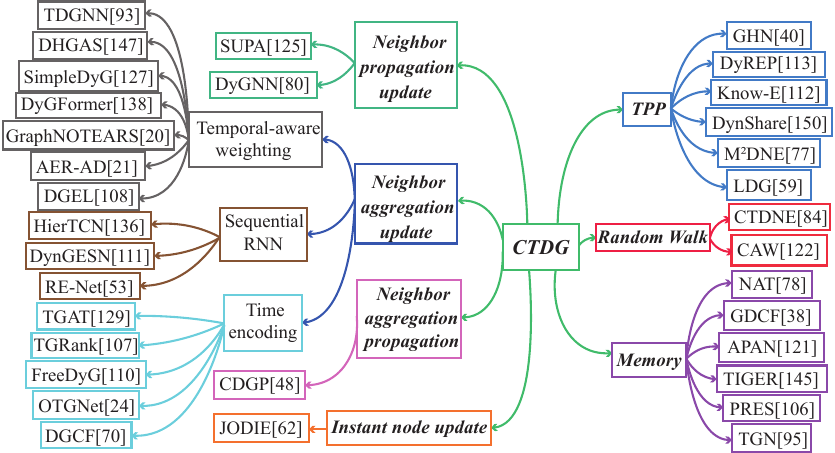}	
	\vspace{-15pt}
	\caption{The taxonomy for CTDG in our survey.} 
	\vspace{-15pt}
	\label{CTDG_photo}%
\end{figure}

\subsection{Continuous-Time Dynamic Graph Models}
\label{subsec:CTDG}

In the realm of CTDG models, notable contributions are categorized into 
\textbf{TPP-based models}, \textbf{memory-based models}, and \textbf{random-walk-based models} 
based on the methods they employ to analyze temporal events. 
For models that do not align with these categories,
we categorize them into \textbf{instant node update model}, \textbf{neighbor aggregation-update model}, \textbf{update-propagation neighbor model}, and \textbf{aggregation-update-propagation model} according to their methods for updating nodes using event streams.
The distinctions between these four node update methods are depicted in Figure~\ref{CTDG_update_photo}.  
The taxonomy of CTDG models is shown in Figure \ref{CTDG_photo}.

\subsubsection{TPP-Based Models}
TPP is a valuable tool for modeling temporal dynamics, as discussed in \S\ref{subsec:learning}.
Several works~\cite{Know-E, DyREP, M2DNE, GHN, LDG, DynShare} utilize TPP in their frameworks. 
Know-E~\cite{Know-E} is pioneering in employing TPP for CTDG and capable of predicting potential event occurrence times. 
It updates node embeddings based on incoming events and recent relationship and time series.  
DyREP~\cite{DyREP} divides model updates into three segments and incorporates TPP to calculate time attention for neighboring nodes. 
$\text{M}^{\text{2}}\text{DNE}$~\cite{M2DNE} introduces macro constraints to scale the network through equations, complemented by micro-level time attention aggregation. 
GHN~\cite{GHN} utilizes the Hawkes process to capture entity interaction time dynamics in TKGs and introduces a novel time knowledge graph connection prediction ranking metric. 
LDG~\cite{LDG} enhances previous approaches by addressing issues related to long-term edge information quality. 
DynShare~\cite{DynShare} designs an interval-aware personalized projection operator using TPP to enable diverse user mode selections within the same time interval.   
While these algorithms all leverage TPP, variations exist in their implementations. 
For instance, DyREP, GHN, LDG, and DynShare directly incorporate TPP in the entire forward propagation calculation, whereas Know-E and $\text{M}^{\text{2}}\text{DNE}$ use TPP for objective function computations.

\subsubsection{Memory-Based Models} 
A part of CTDG models employs a memory mechanism to retain historical information for facilitating embedded updates. 
TGN~\cite{TGN} leverages memory to store historical node data, enabling updates upon the arrival of event streams. 
APAN~\cite{APAN} separates model reasoning from graph computation through an asynchronous processing approach, employing a mailbox mechanism to propagate information to neighbors' mailboxes post-event update, eliminating the need for neighbor aggregation during updates. 
NAT~\cite{NAT} introduces a unique dictionary-based neighborhood representation and N-cache data structure to enable parallel access and updates of these dictionary representations on GPUs. 
GDCF~\cite{GDCF} focuses on crowd flow modeling and employs a memory module to enhance update speed. 
TIGER~\cite{TIGER} introduces a dual memory module to address the limitations of TGN. 
PRES~\cite{PRES} utilizes gaussian mixture models for correction and predicts new memory based on historical time series gradients.

\subsubsection{Random-Walk-Based Models}
Random walk has proven to be effective in graph learning, as discussed in \S\ref{subsec:learning}.  
CTDNE~\cite{CTDNE} is a pioneering algorithm in dynamic graph embedding that incorporates time embeddings into network embeddings.  
This model introduces a comprehensive framework that integrates time dependencies into node embeddings and employs a depth map model through random walks.
CAW~\cite{CAW} represents a temporal network using causal anonymous random walks extracted from temporal random walk, and adopts an anonymization strategy that keep the method inductive and establishes correlation between motifs.

\subsubsection{Instant Node Update Models}
These models update the node embedding based solely on the current event and directly involved nodes, without considering the influence of neighbors, as depicted in Figure~\ref{CTDG_update_photo}(a).  
For instance, JODIE~\cite{JODIE} uses an embedded projection module to predict the future node embedding trajectory.  
It partitions the embedding into dynamic and static components to capture time-varying and time-invariant features, respectively.

\subsubsection{Neighbor Aggregation-Update Models}
These models leverage a combination of event data, historical node information, and neighbor information to update node embeddings, as shown in Figure~\ref{CTDG_update_photo}(b). 
This approach is frequently employed in CTDG models and represents a conventional method for dynamic graph embedding. 
Within this category, classification can be further refined based on the methods used to integrate time information for graph updates:

\noindent\textbf{Temporal-Aware Weighting Models.} 
TDGNN~\cite{TDGNN} introduces a dynamic network representation learning framework that effectively captures node representations by incorporating node characteristics and edge time information using the TDAgg aggregation function. 
DGELL~\cite{DGEL} outlines three key processes (inherent interaction potential, time attenuation neighbor enhancement, and symbiotic local learning) to comprehensively update dynamic node embeddings with rich graph information.
AER-AD~\cite{AER-AD} focuses on inductive anomaly detection in attribute and non-attribute dynamic graphs, utilizing anonymous edge representation for detecting anomalies in dynamic bipartite graphs in an inductive setting.
GraphNOTEARS~\cite{GraphNOTEARS} was the first to study the learning problem of directed acyclic graphs and develop a score-based learning method.
SimpleDyG~\cite{SimpleDyG} proposes a novel strategy that maps dynamic graphs to sequences to enhance scalability.   
DyGFormer~\cite{DyGFormer} presents a transformer-based dynamic graph learning architecture that effectively captures node correlations and long-term temporal dependencies.
DHGAS~\cite{DHGAS} introduces the first dynamic heterogeneous graph neural architecture search method, featuring a unified dynamic heterogeneous graph attention (DHGA) framework that allows each node to focus on its heterogeneity and dynamic neighbors simultaneously.

\noindent\textbf{Sequential RNN Models.} 
RE-Net~\cite{RE-Net} effectively models time, relationships, and interactions between nodes by utilizing RNN to capture the dynamics of time and relationships. 
Its neighborhood aggregation module combines information from neighboring nodes to handle multiple concurrent interactions at the same timestamp.
HierTCN~\cite{HierTCN} employs RNN at the high-level to learn long-term interests, while 
TCN~\cite{TCN} is used at the low-level to predict the next interaction based on long-term interests and short-term interactions.  
DynGESN~\cite{DynGESN} introduces echo state networks (ESN)~\cite{ESN} for dynamic graph modeling.

\noindent\textbf{Time Encoding Models.} 
Models in this category incorporate time information as part of the embedding to influence the calculation process.  
TGAT~\cite{TGAT} uses self-attention and time encoding to create the TGAT layer, achieving similar processing capabilities as GAT on static graphs by stacking TGAT layers.
DGCF~\cite{DGCF} effectively models dynamic user-project relationships, capturing both collaborative and sequential connections.  
OTGNet~\cite{OTGNet} extends TGAT's application to open graphs, enabling handling of open-time dynamic graphs.  
TGRank~\cite{TGRank} boosts the model's link prediction expressiveness, allowing for prediction of crucial structural information.  
FreeDyG~\cite{FreeDyG} introduces a node interaction frequency encoding module that explicitly captures common neighbor proportions and node pair interaction frequencies to address the {\em shift} phenomenon.  

\subsubsection{Update-Propagation Neighbor Models}
These models only utilize event and historical node information to update, embed, and propagate data to neighbors, as depicted in Figure~\ref{CTDG_update_photo}(c).  
In DyGNN~\cite{DyGNN}, the nodes engaged in the event are initially updated using LSTM, followed by updating the neighbors of these nodes with event information.  
SUPA~\cite{SUPA} generates a sample graph for the event-involved nodes, updates them with event flow data, and propagates the updated information throughout the sample graph.

\subsubsection{Aggregation-Update-Propagation Models}
These models rely on historical neighbor information for embedding updates, followed by propagating the updated information to neighbors, as illustrated in Figure~\ref{CTDG_update_photo}(d).  
An example is CDGP~\cite{CDGP}, a popularity prediction model on community dynamic graphs. 
CDGP identifies the community for the event, aggregates nodes within the community for embedding updates, and extends the community's influence to other nodes within the same community.

\begin{table*}[t]
    \caption{The comparison between the existing dynamic GNN training frameworks}
    \vspace{-10pt}
    \renewcommand{\arraystretch}{1.5}
    \scalebox{0.75}{
    \begin{tabular}{ | c | c| c| c |c |c |c |c |c |} 
     \hline
     \multirow{2}{*}{\diagbox{Systems}{Features}} & \multicolumn{2}{c|}{Universality} & \multicolumn{3}{c|}{Expandability} &  \multicolumn{3}{c|}{Supported functionalities} \\
      \cline{2-9}
      & DTDG & CTDG & \thead{Single-machine \\ single-GPU} & \thead{Single-machine \\ multi-GPU} & \thead{Multi-machine \\ multi-GPU} & Temporal neighbor storage & Feature extraction optimizing & Temporal parallel training \\
     \hline
      PyGT~\cite{PYGT} & \Checkmark & \XSolid & \Checkmark & \XSolid & \XSolid & Store snapshots & \XSolid & \XSolid \\
      \hline
      ESDG~\cite{ESDG} & \Checkmark & \XSolid & \Checkmark & \Checkmark & \Checkmark & Store snapshots & \XSolid & Snapshot parallelism \\
      \hline
      PiPAD~\cite{PiPAD} & \Checkmark & \XSolid & \Checkmark & \Checkmark & \XSolid & Store snapshots & Extract common snapshots & Snapshot parallelism \\
      \hline
      DynaGraph~\cite{DynaGraph} & \Checkmark & \XSolid & \Checkmark & \Checkmark & \Checkmark & Store snapshots & \XSolid & Snapshot partition parallel \\
      \hline
      BLAD~\cite{BLAD} & \Checkmark & \XSolid & \Checkmark & \Checkmark & \Checkmark & Store snapshots & \XSolid & Operator parallelism \\
      \hline
      TGL~\cite{TGL} & \Checkmark & \Checkmark & \Checkmark & \Checkmark & \XSolid & Neighbor static sorting & \XSolid & Small-batch parallelism \\
      \hline
      DistTGL~\cite{DistTGL} & \Checkmark & \Checkmark & \Checkmark & \Checkmark & \Checkmark & Neighbor static sorting & \XSolid & Memory parallelism \\
      \hline
      NeutronStream~\cite{NeutronStream} & \XSolid & \Checkmark & \Checkmark & \XSolid & \XSolid & Store event stream & \XSolid & Event group parallelism \\
      \hline
      SPEED~\cite{SPEED} & \XSolid & \Checkmark & \Checkmark & \Checkmark & \XSolid & Store static graph & \XSolid & Subgraph partition parallel \\
      \hline
      DyGLib~\cite{DyGFormer} & \XSolid & \Checkmark & \Checkmark & \XSolid & \XSolid & Store event stream & \XSolid & \XSolid \\
      \hline
      Zebra~\cite{Zebra} & \XSolid & \Checkmark & \Checkmark & \XSolid & \XSolid & Store event stream & \XSolid & \XSolid \\
      \hline
      GNNFlow~\cite{GNNFlow} & \XSolid & \Checkmark & \Checkmark & \Checkmark & \Checkmark & Store event stream & Vectorized cache & \XSolid \\
     \hline
    \end{tabular}
    }
    \vspace{-5pt}
\label{table:DGNN_framework_taxonomy}
\end{table*}

\section{Dynamic GNN Training Frameworks}
\label{sec:frameworks}

In this section, we offer a thorough survey of the latest dynamic GNN frameworks, encompassing 5 discrete-time dynamic graph (DTDG) frameworks and 7 continuous-time dynamic graph (CTDG) frameworks.  
We start by outlining the requirements for dynamic GNN frameworks, then delve into the existing DTDG frameworks and CTDG frameworks, respectively. 

\subsection{Needs for DGNN Frameworks.}
\label{subsec:framework_needs}

While research on dynamic GNN models has advanced significantly, the focus has been primarily on enhancing model training accuracy and expanding application domains.  
The high computational complexity of these models often results in low training efficiency and limited scalability on large graphs, making them suitable mainly for small graphs within thousands of vertices and tens of thousands of edges.  
This limitation hinders their practical application.  
Hence, there is a critical requirement for a framework that enhances the training efficiency and scalability of dynamic GNNs on large-scale dynamic graphs by optimizing system execution efficiency.

Recently, specialized GNN training frameworks like PyG~\cite{PyG} and DGL~\cite{DGL} have been developed to simplify programming and enhance training efficiency for various GNN models.  
These frameworks introduce optimization strategies on top of fundamental training modules.  
For instance, NeuGraph~\cite{NeuGraph} and Roc~\cite{Roc} optimize vertex access and load balancing through improved graph partitioning strategies.  
AliGraph~\cite{AliGraph} and DUCATI~\cite{DUCATI} reduce data transmission by caching embeddings and intermediate training results.  
P3~\cite{P3} and PipeGCN~\cite{PipeGCN} employ parallel training strategies to accelerate GNN training.  
However, these frameworks primarily cater to training static graphs and lack essential mechanisms for updating graph structures and modeling temporal information in dynamic GNN models.  
Key modules like dynamic graph loading, temporal neighbor sampling, and temporal message passing are missing.  
The temporal dependency of dynamic graphs also complicates parallel computing.  
Therefore, users must develop these modules themselves to construct dynamic GNN models and carefully handle dynamic graphs to ensure the temporal consistency of data training, resulting in heavy development costs for users. 

In the past two years, several dynamic GNN training frameworks have emerged, 
aiming to better support the training of dynamic GNN models, including both discrete-time and continuous-time dynamic graph models. 
However, due to the differences in training processes between these two types of models, most frameworks are optimized for only one type of models. 
We conducted a study on the 12 existing dynamic GNN training frameworks and compiled a detailed comparison of their universality, expandability, and supported functionalities, as outlined in Table~\ref{table:DGNN_framework_taxonomy}.

\subsection{Discrete-Time DGNN Frameworks}
\label{subsec:DTDG_frameworks}

Discrete-time DGNNs typically treat the dynamic graph as a sequence of graph snapshots.
PyGT~\cite{PYGT}, an extension of PyG, introduces three snapshot-based dynamic graph types and provides a unified data loader for each type of dynamic graph to support for training DTDG models. 
However, it is limited to full-batch training on single GPU for small-scale graphs. 
ESDG~\cite{ESDG} utilizes snapshot partitioning to distribute multiple graph snapshots to different devices for parallel GNN training, and after finishing the GNN computation, redistributes them to colocate the same vertices from different snapshots on the same device for parallel RNN computation. 
PiPAD~\cite{PiPAD}, based on the similarity of the structures between adjacent graph snapshots, reduces data transmission by transmitting the common parts of a set of consecutive graph snapshots and the individual parts of each graph snapshot, ensuring that these snapshots can be computed in parallel on GPU. 
DynaGraph~\cite{DynaGraph} proposes a time fusion mechanism to concatenate node features of multiple snapshots in a single graph convolution operation to fully utilize the GPU resource.
BLAD~\cite{BLAD} explores fine-grained operator-level parallelism to fully leverage GPU resources by simultaneously executing memory-intensive graph operators and compute-intensive neural network operators on a single GPU.

\vspace{-10pt}
\subsection{Continuous-Time DGNN Frameworks}
\label{subsec:CTDG_frameworks}

Traditional static graph storage formats like compressed sparse row (CSR) formats are not well-suited for dynamic graphs as they struggle to efficiently handle edge and vertex insertions and deletions. 
Dynamic graph storage also needs to handle temporal graph sampling effectively, considering only edges up to the current timestamp.  
TGL~\cite{TGL} addresses these challenges by introducing a framework that abstracts key components for training dynamic GNN models, including temporal sampler, temporal message mailbox, vertex history memory, memory updater, and message passing engine. 
It also utilizes sorted static neighbor storage format T-CSR for dynamic graph storage, 
and parallel samplers and random block scheduling techniques for enhancing training efficiency.  
GNNFlow~\cite{GNNFlow} proposes a time-indexed block-based data structure for dynamic graph storage to facilitate edge sampling and updates.  
It also implements caching mechanisms by specifically caching frequently used edge features to optimize CPU-GPU data transfers performance.

CTDG models view dynamic graphs as sequences of events with temporal dependencies, making parallelization challenging.  
TGL uses mini-batch parallelism but overlooks dependencies within individual mini-batches, potentially impact training accuracy. 
NeutronStream~\cite{NeutronStream} employs an adaptive sliding window training approach to capture temporal dependencies and identify parallelizable event groups while maintaining temporal order.  
DistTGL~\cite{DistTGL} extends TGL by introducing distributed training methods, such as epoch parallelism and memory parallelism, to enhance scalability across multiple GPUs. 
SPEED~\cite{SPEED} customizes partitioning strategies for parallel training to ensure load balance and minimize replication factors while preserving temporal information.  
DyGLib~\cite{DyGFormer} offers a comprehensive library for dynamic graph learning, featuring standardized training processes, scalable coding interfaces, and thorough evaluation protocols. 
Zebra~\cite{Zebra} introduces T-PPR by combining random walks with dynamic GNNs to quickly generate node embeddings through top-k T-PPR queries.

\section{Dynamic GNN Benchmarks}
\label{sec:benchmarks}

\subsection{Datasets}
\label{subsec:datasets}

Table~\ref{table:datasets} shows a summary of commonly used graph datasets in dynamic GNN evaluation. 
It includes details on application domains, node and edge counts and dimensions, timestamp ranges, label information, and dataset sizes. 
The datasets highlighted in bold have been chosen for comparison evaluation (\S\ref{sec:experiments}). 
Subsequently, we provide a brief introduction to these datasets.

\begin{table}[!t]
  \caption{Summary of common graph datasets, 
  where $|V|$ and $|E|$ stand for the number of nodes and edges, 
  $Range(t)$ denotes the timestamp range, and $Sizes$ specifies the total file sizes. 
  }
  \label{table:datasets}
  \vspace{-10pt}
  \scalebox{0.8}{
  \begin{tabular}{l c c c c} 
    \hline
    Datasets &  $|V|$ & $|E|$ & $Range(t)$ & $Sizes$ \\ 
    \hline
    {\bf Reddit}~\cite{jodie_data} & 10984 & 672447  & 0$\sim$2678390 & 2.2GB \\ 
    {\bf DGraphFin}~\cite{DGraphFin_data}  & 4889537 & 4300999 & 1$\sim$821 & 649MB  \\ 
    Enron~\cite{Enron_Flights_data} & 184 & 125,235 & 0$\sim$113740399 & 35MB \\ 
    Facebook~\cite{facebook_data} & 63731 & 1269502 & 1157454929$\sim$1232576125 & 26MB \\ 
    Social Evolution~\cite{Social_Evolution_data} & 74 & 2,099,520 & 1188972131$\sim$1247740843  & 148MB \\ 
    UCI~\cite{UCI_data} & 899 & 33720 & 1084560796$\sim$1098772901 & 668KB \\ 
    {\bf Wikipedia}~\cite{jodie_data}  & 9227 & 157474 & 0$\sim$2678373 & 534MB \\ 
    {\bf MOOC}~\cite{jodie_data} & 7144 & 411749 & 0$\sim$2572086 & 40MB \\ 
    {\bf ML25M}~\cite{ML25M_data} & 221588 & 25000095  & 789652009$\sim$1574327703  & 647MB  \\ 
    LastFM~\cite{jodie_data}  & 1980 & 1293103  & 0$\sim$137107267  & 37MB \\ 
    FB-FORUM~\cite{FB-FORUM_data}   & 899 & 33720  & 1084585996$\sim$1098798101  & 612KB \\ 
    DBLP~\cite{DBLP_data}   & 28,086 & 162,451  & 1$\sim$27  & 375MB \\ 
    Yelp~\cite{Yelp_data}  & 2138275 & 6990280 & 1108495402$\sim$1642592925 & 5.0GB  \\  
    ICEWS05-15~\cite{ICEWS_data} & 10,094 & 461,329  & 1104537600$\sim$1451520000  & 30MB  \\ 
    ICEWS14~\cite{ICEWS_data}  & 6,869 & 96,730 & 1388534400$\sim$1419984000  & 6MB \\ 
    ICEWS18~\cite{icews18_data}  & 23,033 & 741820  & 1514764800$\sim$1546214400  & 184MB  \\ 
    GDELT~\cite{GDELT_data}  & 16682 & 191290882 & 0$\sim$175283  & 82.3GB \\ 
    Bitcoin-OTC~\cite{Bitcoin-OTC_Alpha_as_data} & 5881 & 35592  & 1289241942$\sim$1453684324  & 988KB \\ 
    Bitcoin-Alpha~\cite{Bitcoin-OTC_Alpha_as_data} & 3782 & 24186  & 1289192400$\sim$1453684324  & 492KB \\ 
    AS-733~\cite{Bitcoin-OTC_Alpha_as_data} & 7716 & 11965533  & 939340800$\sim$946771200 & 115MB \\ 
    {\bf Flights}~\cite{Enron_Flights_data} & 13169 & 1927145  & 0$\sim$121 & 32MB \\ 
    \hline
  \end{tabular}
  }
  \vspace{-15pt}
  \end{table}

\subsubsection{Social Networks} 
\textbf{Reddit}~\cite{JODIE} consists of one month of user posts on subreddits, with users and subreddits as nodes and timestamped posting requests as links. 
\textbf{DGraphFin}~\cite{dgraphfin} is the real social network in financial industry provided by Finvolution Group, with nodes representing Finvolution users and edges indicating emergency contacts. 
\textbf{Enron}~\cite{enron} is an mail dataset capturing interactions among Enron Inc. employees, where communication links denote email exchanges among core employees. 
\textbf{Facebook}~\cite{facebook} is a user interaction network where nodes represent users and edges signify interactions between users. 
\textbf{Social Evolution}~\cite{social-evolution} comes from the MIT Human Dynamics Lab 
focusing on social relationship evolution and the degree of closeness between individuals.
\textbf{UCI}~\cite{UCI} is an online community of students from the University of California, Irvine, with links representing messages exchanged between users.

\subsubsection{Interaction Networks} 
\textbf{Wikipedia}~\cite{JODIE} is a bipartite interaction graph capturing user edits on Wikipedia pages over a month, with nodes representing users and pages, and links representing editing behaviors. 
\textbf{MOOC}~\cite{JODIE} records student behaviors in Massive Open Online Courses (MOOCs), such as watching videos and submitting answers. 
\textbf{ML25M}~\cite{ml25m} contains user ratings of movies, reflecting their preferences and levels of interest in various movies.
\textbf{LastFM}~\cite{JODIE}provides information on user-listened songs over a month. 
\textbf{FB-FORUM}~\cite{FB-forum}is the Facebook-like forum network from the same online community as the social network dataset, focusing on user activities in the forum. 
\textbf{DBLP}~\cite{DBLP} captures academic collaboration network dataset with nodes representing authors and edges denoting co-authored papers. 
\textbf{Yelp}~\cite{DySAT} is a large business review website where users can upload comments to review businesses and discover interested ones based on others' comments.

\subsubsection{Event Networks} 
\textbf{ICEWS}~\cite{ICEWS05-15,ICEWS14/18} and \textbf{GDELT}~\cite{gdelt} are event networks 
where nodes represent actors and temporal edges represent point-time events derived from news and articles, such as "yield," "make public statement," "threaten," and more, providing a dynamic view of real-world interactions and occurrences.

\subsubsection{Trade Networks.} 
\textbf{Bitcoin-OTC}~\cite{bitcoinotc} and \textbf{Bitcoin-Alpha}~\cite{bitcoinalpha} are who-trusts-whom networks of bitcoin users trading on the bitcoin-otc platform and btc-alpha platform, respectively. 

\subsubsection{Traffic Networks.} 
\textbf{AS-733}~\cite{AS-733} is a communication network composed of routers used to exchange traffic with peers. 
\textbf{Flights}~\cite{flights} is a transportation map where nodes represent airports and edges represent flights between these airports.

\begin{table}[!t]
  \caption{Common evaluation metrics in dynamic GNNs.}
  \label{table:metrics}
  \vspace{-10pt}
  \renewcommand{\arraystretch}{1.8}
  \scalebox{0.8}{
  \begin{tabular}{| c | c | c |} 
   \hline
   Applications & Metrics & Equations \\ 
   \hline
    \multirow{5}{*}{\thead{Binary\\classification}} & Precision & $\frac{TP}{TP+FP}$ \\ 
    \cline{2-3}
     & Accuracy & $\frac{TP+TN}{TP+FP+FN+TN}$ \\ 
     \cline{2-3}
     & Recall & $\frac{TP}{TP+FN}$ \\ 
     \cline{2-3}
     & F1 score & $\frac{2\times Precision \times Recall}{Precision+Recall}$ \\ 
     \cline{2-3}
     & AUC  & $TPR=\frac{TP}{TP+FN},\,FPR=\frac{FP}{FP+TN}$ \\ 
   \hline
      \multirow{6}{*}{\thead{Multi\\classification}} & \multirow{3}{*}{Micro-F1}  & $(1)Precision_{micro}=\frac{\sum_{i}TP_i}{\sum_{i}TP_i+FP_i}$ \\
       &  & $(2)Recall_{micro}=\frac{\sum_{i}TP_i}{\sum_{i}TP_i+FN_i}$ \\
       &  & $\frac{2\times Precision_{micro} \times Recall_{micro}}{Precision_{micro}+Recall_{micro}}$ \\
      \cline{2-3}
       & \multirow{3}{*}{Macro-F1} & $(1)Precison_{macro}=\frac{1}{N}\sum_{i}\frac{TP_i}{TP_i+FP_i}$ \\ 
       &  & $(2)Recall_{macro}=\frac{1}{N}\sum_{i}\frac{TP_i}{TP_i+FN_i}$ \\
       &  & $\frac{2\times Precison_{macro}\times Recall_{macro}}{Precison_{macro}+Recall_{macro}}$ \\
    \hline
      \multirow{6}{*}{\thead{Recommendation\\system}} & Precision@K & $\frac{TP@K}{TP@k+FP@K}$ \\  
      \cline{2-3}
       & Recall@K  & $\frac{TP@K}{TP@k+FN@K}$ \\ 
       \cline{2-3}
      & MR & $\frac{1}{|S|}\sum_{i=1}^{|S|}{rank_i}$ \\
      \cline{2-3}
      & MRR & $\frac{1}{|S|}\sum_{i=1}^{|S|}\frac{1}{rank_i}$ \\
      \cline{2-3}
      & HITS@K & $\frac{1}{|S|}\sum_{i=1}^{|S|}IF(rank_i\leq k)$ \\
     \hline
       \multirow{3}{*}{\thead{Regression\\Model\\Performance}} & RMSE  & $\sqrt{\frac{\sum_{i=1}^{n}(y_i-\hat{y}_i)^2}{n}}$ \\
       \cline{2-3}
        & MAE & $\frac{\sum_{i=1}^{n}|y_i-\hat{y}_i|}{n}$ \\
        \cline{2-3}
        & MAPE & $\frac{\sum_{i=1}^{n}|\frac{y_i-\hat{y}_i}{y_i}|}{n}$ \\
    \hline
  \end{tabular}
  }
  \vspace{-10pt}
\end{table}

\begin{table}[!t]
  \centering
  \caption{Confusion matrix in binary classification tasks.}
  \vspace{-10pt}
  \scalebox{0.8}{
  \begin{tabular}{c|c|c|c}
      \hline
     \multicolumn{2}{c|}{}  & \multicolumn{2}{c}{Prediction}  \\
     \cline{3-4}
    \multicolumn{2}{c|}{} & Positive & Negative \\
     \hline
     \multirow{2}{*}{Actual} & True & True Positive(TP) & False Negative(FN) \\
     \cline{2-4}
     & False & False Positive(FP) & True Negative(TN)\\
     \hline
  \end{tabular}
  }
  \vspace{-10pt}
  \label{table:confusion_matrix}
\end{table}

\subsection{Metrics}
\label{subsec:metrics}

Here is a brief overview of commonly used evaluation metrics to showcase the performance of different dynamic GNN models and frameworks.  
For detailed calculation equations, refer to Table~\ref{table:metrics}.

\subsubsection{Binary Classification Performance}
The confusion matrix for binary classification tasks is shown in Table~\ref{table:confusion_matrix}, where rows represent predicted classes and columns represent actual classes. 
\textbf{Precision} measures the proportion of correctly predicted positive samples among all predicted positives. 
\textbf{Accuracy} indicates the proportion of correctly predicted samples out of all samples.
\textbf{Recall} assesses the proportion of actual positive samples correctly predicted. 
\textbf{F1 Score} is the harmonic mean of precision and recall, offering a balanced measure of classification performance for both positive and negative cases. 
\textbf{AUC} (Area under the curve) typically denotes the area under the ROC curve, a tool for assessing binary classification models using True Positive Rate (TPR) and False Positive Rate (FPR).

\subsubsection{Multi Classification Performance}
In multi-class classification tasks, key performance metrics include:
\textbf{Micro-F1 Score} computes the F1 score considering total true positives, false positives, and false negatives across all categories. 
\textbf{Macro-F1 Score} computes the average of F1 scores for individual categories. 
\textbf{Micro-AUC} is calculated by micro-averaging true positive rates and false positive rates for all categories. 
\textbf{Macro-AUC} is obtained by macro-averaging AUC values for each category.

\subsubsection{Recommendation System Performance}
In recommendation system tasks, 
\textbf{Precision@K} measures the proportion of the top $K$ recommended items to the total number of recommended items.
\textbf{Recall@K} measures the proportion of the top $k$ recommended items to the total number of relevant items.
\textbf{MR} (Mean Rank) measures the average recommended ranking $rank_i$ of all users $S$.
\textbf{MRR} (Mean Reciprocal Rank) measures the average reciprocal of the recommended ranking of all users.
\textbf{HITS@K} measures the accuracy of the top K recommendation results by assessing how many of them align with users' genuine interests.
Function IF(·) outputs 1 if the condition is true and 0 otherwise. 

\subsubsection{Regression Model Performance}
In recommendation systems, 
\textbf{RMSE} (Root Mean Squared Error) quantifies the difference between predicted and actual values, while
\textbf{MAE} (Mean Absolute Error) calculates the average error between predicted and actual values.
\textbf{MAPE} (Mean Absolute Percentage Error) measures the average percentage difference between predicted and actual values. 

\subsubsection{Efficiency and Scalability}
\textbf{Throughput} measures the number of tasks or data processed within a specific time frame. In deep learning, it typically refers to the number of samples that the model can handle per unit time, providing insights into processing speed.
\textbf{Training time} refers to the duration time the model spends in the training phase. 
\textbf{GPU memory usage} indicates the amount of memory occupied during training or inference on a GPU. Efficient memory management is crucial for optimizing model performance and scalability.
\textbf{Parameter size} quantifies the total number of learnable parameters in the model. 
The parameter size directly impacts the model's complexity, storage requirements, and computational efficiency, influencing scalability and performance. 

\section{Experimental Comparison}
\label{sec:experiments}

We thoroughly evaluate various dynamic GNN models and frameworks in terms of training accuracy, efficiency, and memory usage. 
We first compare several classic DTDG and CTDG GNN models (\S\ref{model_evaluation}), 
then compare models implemented on optimized frameworks (\S\ref{compare_models_and_frameworks}), analyze multi-GPU scalability (\S\ref{scalability}). 

\subsection{Experiment Setting}
\label{experiment_setting}

\noindent\textbf{Test Setup:} 
We perform our experiments on two Ubuntu 22.04.3 LTS machines,  
each featuring an Intel Xeon Gold 6342 CPU @ 2.80GHz and four Nvidia A40 48GB GPUs.  These machines are equipped with 1TB of memory and 40TB of disk space.

\noindent\textbf{Baseline Models and Frameworks:} 
We compare six CTDG GNN models (JODIE, TGAT, TGN, APAN, CAW, DyREP) 
and three DTDG GNN models (EvolveGCN, Roland, DySAT), 
along with three dynamic GNN frameworks (TGL, DistTGL, SPEED). 
For JODIE and DyREP models, we use the implementation provided by TGN.  
We utilize the most efficient version, TGN-attn, for the TGN model.  
EvolveGCN have two versions EvolveGCN-H and EvolveGCN-O, using GRU and LSTM as components of learning time, respectively.
All CTDG models are set by default with 1 GNN layer and a batch size of 1000, and follow the same early stopping condition (3 epochs) as AP for consistency.
For DTDG models, we maintain uniform snapshot intervals across datasets.  
Remaining parameters are left at default values as they are typically optimal.

\noindent\textbf{Datasets and Metrics:} 
We evaluate on six graph datasets (Wikipedia, Reddit, MOOC, Flights, ML25M, and DGraphFin).  
Specific details of the datasets can be found in Table~\ref{table:datasets}.  
In evaluating training accuracy, we consider four key metrics (AUC, AP, Recall, and Accuracy).  
Furthermore, we analyze time cost and memory usage to assess training efficiency and scalability.
Our primary focus is on the link prediction task, the most common task for dynamic GNNs.  

\begin{table*}[!t]
    \caption{Converge accuracy on dynamic GNN models for link prediction task.}
    \vspace{-10pt}
    \renewcommand{\arraystretch}{1.2}
    \scalebox{0.7}{
    \begin{tabular}{| c | c | c | c | c | c | c | c | c | c | c | c | c | c | c | c | c | c |} 
     \hline
     \multirow{2}{*}{Type} & \multirow{2}{*}{Model} & \multicolumn{4}{c|}{Wikipedia} & \multicolumn{4}{c|}{Reddit}  & \multicolumn{4}{c|}{MOOC}  &\multicolumn{4}{c|}{Flights}  \\
     \cline{3-18}
      & & AUC & AP & Recall & Accuracy & AUC & AP & Recall & Accuracy & AUC & AP & Recall & Accuracy & AUC & AP & Recall & Accuracy  \\
     \hline
     \multirow{6}{*}{CTDG} & JODIE & 0.9325 & 0.9313 & 0.8657 & 0.8515 & 0.9636	&0.9610	&0.9155	&0.9019 &0.6243	&0.5920	&0.6805	&0.5942 & 0.9116	&0.9049	&0.9395	&0.7862  \\
     \cline{2-18}
      &	TGAT &	0.8202 & 0.8339 & 0.7838 & 0.7351 & 0.9585 & 0.9610 & 0.9151 & 0.8877 &0.5637	&0.5720	&0.8419	&0.5460 & 0.7008 & 0.6800 & 0.6589 & 0.6373 \\
     \cline{2-18}
      &	TGN-attn & 0.9781	&0.9788&	0.9076	&0.9191 & 0.9861	&0.9862	&0.9450	&0.9410 &0.8288	&0.8078	&0.6715	&0.7403& \textbf{0.9763} &	\textbf{0.9722}	&0.9833	&0.8927  \\
     \cline{2-18}
      &	APAN & 0.9650 & 0.9573 & \textbf{0.9579} & 0.9094 & \textbf{0.9969} & \textbf{0.9965} & \textbf{0.9886} & \textbf{0.9777} & \textbf{0.9935} & \textbf{0.9883} & 0.9896 & \textbf{0.9859} & 0.8995 & 0.8904 & 0.9411 & 0.7818 \\
     \cline{2-18}
      &	CAW & \textbf{0.9819} & \textbf{0.9854} & 0.9151 & \textbf{0.9430} & 0.9835 & 0.9859 & 0.9210 & 0.9393 &0.7424	&0.7724	&0.8799	&0.6218& 0.9626 & 0.9663 & 0.8843 & \textbf{0.9053}  \\
     \cline{2-18}
      &	DyRep &	0.9171	&0.9214	&0.8195	&0.8293 & 0.9667	&0.9654	&0.8934	&0.9067 &0.6997	&0.6505	&0.4325	&0.6007& 0.9214	&0.9109	& \textbf{0.9842}	&0.7094  \\
     \hline
     \multirow{4}{*}{DTDG} & EvolveGCN-H &	0.9053&	0.6253&	0.7899&	0.8775&	0.7345	&0.4938	&0.4803	&0.8997	&0.7255	&0.5561	&\textbf{0.9951}	&0.2906	&0.9044	&0.5941	&0.9033	&0.7053 \\
     \cline{2-18}
      & EvolveGCN-O & 0.8820&	0.5667	&0.7242	&0.8832	&0.9107	&0.6490	&0.9101	&0.6945	&0.9366	&0.8214	&0.9432	&0.7964	&0.8775& 0.3787	&0.8647	&0.7272 \\
     \cline{2-18}
      &	Roland & 0.8584	&0.8748	&0.7473	&0.8145 & 0.9329	&0.9511	&0.7894	&0.8454&  0.9409	&0.9571	&0.7981	&0.7582 & 0.8512	&0.8599	&0.3640	&0.5168 \\
     \cline{2-18}
      &	DySAT &	0.8668&	0.8987	&0.9293	&0.6513	&0.9179	&0.9345	&0.9471	&0.7195	&0.4551	&0.4336	&0.5596	&0.4862 &\multicolumn{4}{c|}{OOM} \\
     \hline
    \end{tabular}
    }
    \label{link_prediction_task_result}
	\vspace{-10pt}
\end{table*}

\begin{figure}
	\centering 
	\hspace{-15pt}
    \includegraphics[width=0.5\textwidth, angle=0]{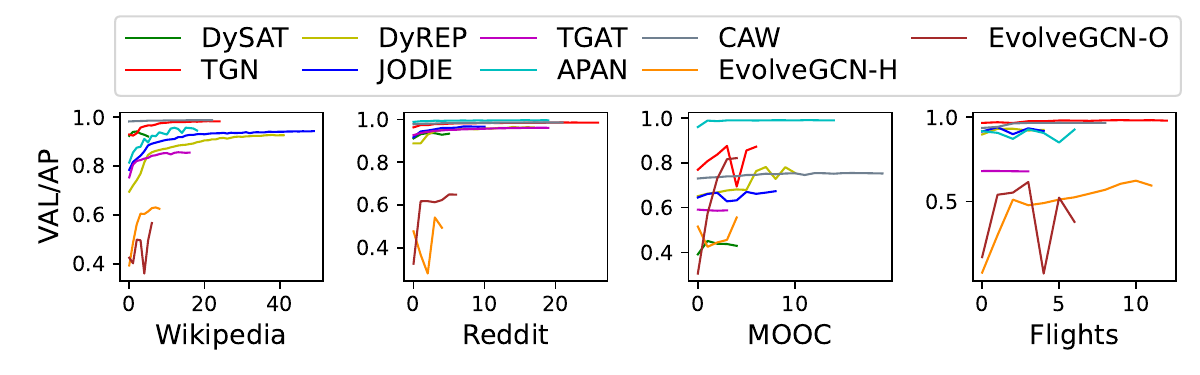}	
    \vspace{-20pt}
	\caption{Val.ap after each training epoch.} 
	\vspace{-15pt}
	\label{experiment:convergence_ap}%
\end{figure}
\begin{figure}
	\centering 
	\hspace{-15pt}
    \includegraphics[width=0.5\textwidth, angle=0]{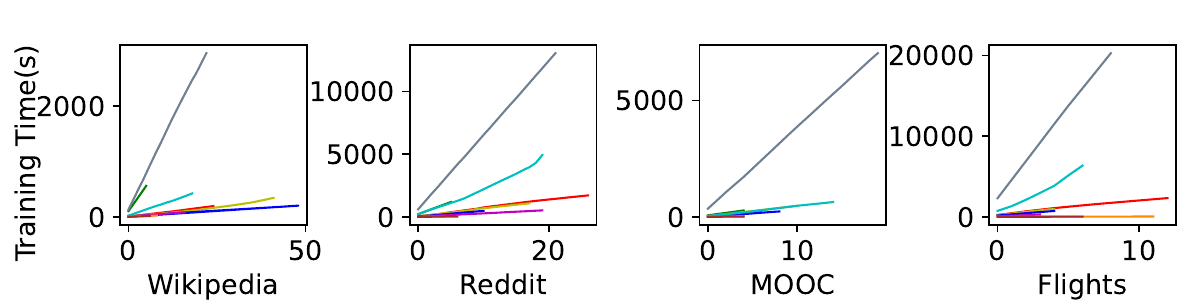}	
    \vspace{-15pt}
	\caption{Total training time cost after each epoch.} 
	\vspace{-10pt}
	\label{experiment:convergence_time}%
\end{figure}

\begin{figure}
	\centering 
	\includegraphics[width=0.46\textwidth, angle=0]{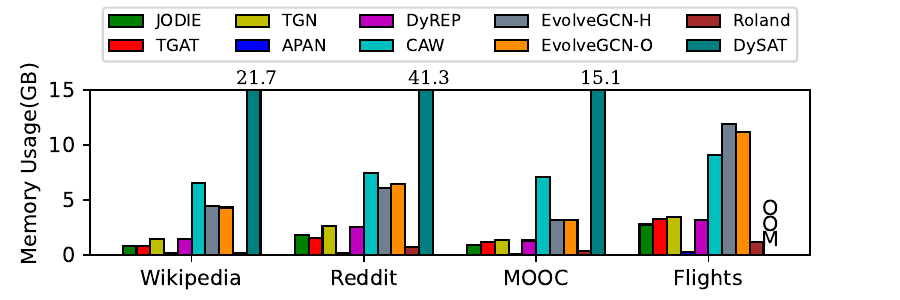}	
    \vspace{-15pt}
	\caption{Memory usage of training dynamic GNN models.}
    \vspace{-15pt}
	\label{experiment:model_memory_use}%
\end{figure}

\subsection{Comparison of Dynamic GNN Models}
\label{model_evaluation} 

We extensively compared the nine dynamic GNN models by training them to convergence. 
The convergence accuracy are detailed in Table~\ref{link_prediction_task_result}. 
Notably, most models encountered out-of-memory (OOM) issues on the DGraphFin and ML25M datasets, except for APAN and CAW. 
However, even APAN and CAW exhibited significant slowness and failed to complete within 48 hours. 
Hence, the results on these two datasets are excluded. 
Furthermore, we illustrate the convergence curves of validated AP and time cost after each epoch in Figure \ref{experiment:convergence_ap} and \ref{experiment:convergence_time}, respectively.
Roland's distinctive training method allows it to converge after just one epoch, hence its results are not included in these figures. 
GPU memory usage is shown in Figure~\ref{experiment:model_memory_use}. 

Table~\ref{link_prediction_task_result} highlights the consistent top performance of CTDG models, with TGN, APAN, CAW, and DyREP excelling on various datasets and metrics. 
CAW stands out on the Wikipedia dataset, showcasing superior performance across all metrics due to its unique anonymous causal walk approach. 
APAN shines on the Reddit and MOOC datasets, demonstrating versatility beyond financial scenarios into user interaction datasets. 
TGN excels in AUC/AP on the Flights dataset. 
We also observed that on the MOOC dataset, EvolveGCN-H has a high Recall value but low Accuracy. 
This discrepancy is attributed to a high number of false positives (FP),
indicating that the model has good discrimination for positive samples, but cannot make correct judgments for negative samples.
On the other hand, analysis from Figure~\ref{experiment:convergence_time} and Figure~\ref{experiment:model_memory_use} shows that the high training accuracy of CAW and APAN comes with significant time and memory costs, especially for CAW. 
TGN maintains commendable performance with acceptable training times. 
DySAT and EvolveGCN exhibit lower accuracy metrics while consume considerable memory, making them less efficient. 
Figure~\ref{experiment:convergence_ap} indicates that higher validation metrics usually lead to superior test results.

\subsection{Comparison of Models on Frameworks}
\label{compare_models_and_frameworks}

In this section, we evaluate these models trained on dedicated optimized dynamic GNN frameworks with open-sources: TGL, DistTGL, and SPEED. 
TGL implemented JODIE, TGAT, TGN, APAN, and DySAT models, while DistTGL only focuses on the TGN model. 
SPEED covers TGN, TGAT, JODIE, DyREP, and TIGE models.  
For TGL's DySAT implementation (TGL-DySAT), we maintain the same time interval size as the model for each dataset. 
We set $top_k=10$ for SPEED to explore potentially higher evaluation metrics.

\begin{figure*}
	\centering 
	\hspace{-0.5cm}\includegraphics[width=1\textwidth, angle=0]{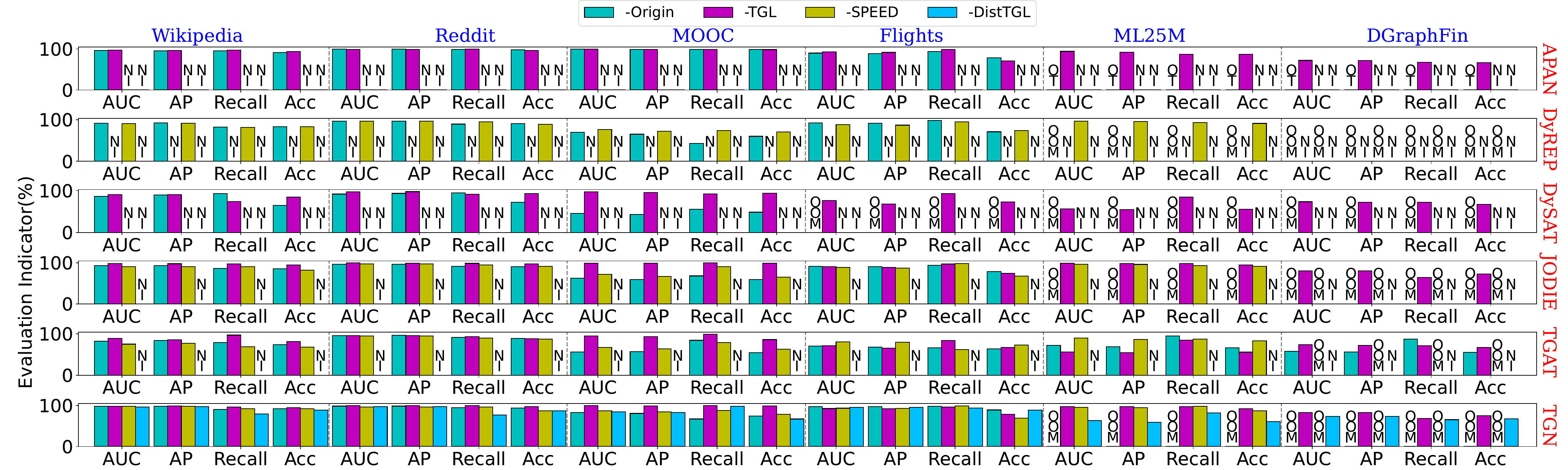}	
    \vspace{-10pt}
	\caption{Comparison of accuracy metrics of training models to converge on frameworks.``Acc'' means ``Accuracy'', ``OT'' means ``OverTime'', ``NI'' means ``Not Implemented'', ``OOM'' meas ``Out of Memory''.} 
	\vspace{-5pt}
	\label{compare_all_indicator}%
\end{figure*}

\noindent\textbf{Accuracy Metrics Comparison:}  
We show the four accuracy metrics of models trained to converge on frameworks in Figure~\ref{compare_all_indicator}. 
Situations facing issues such as not-implemented (NI), out-of-memory (OOM), or overtime (OT) are indicated in the corresponding bar. 
In general, frameworks tend to achieve better accuracy than the origin models, although the model outperforms in some cases.
For example, JODIE on the Wikipedia dataset shows superior AUC/AP/Accuracy results compared to the SPEED framework. 
Frameworks have an advantage over models in their ability to handle large datasets effectively. 
Models often struggle with training on large datasets, whereas frameworks are tailored to support such scenarios efficiently. 
Among TGL, SPEED, and DistTGL comparisons, TGL frequently demonstrates higher accuracy.

\begin{figure*}
	\centering 
	\hspace{-0.5cm}\includegraphics[width=1\textwidth, angle=0]{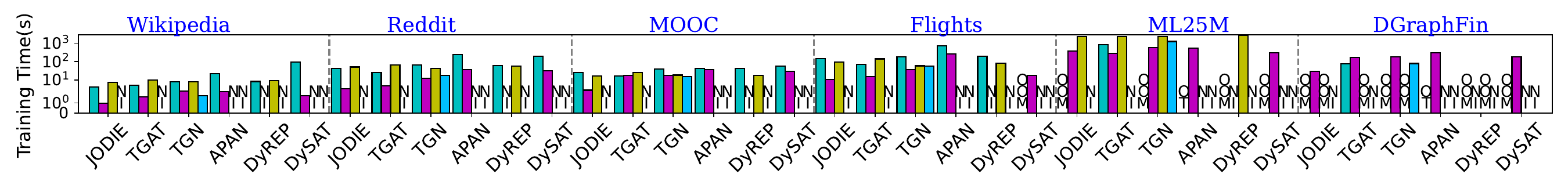}	
    \vspace{-10pt}
	\caption{Comparison of average per epoch training time of models and frameworks, and the y-axis are shown in log. } 
	\vspace{-5pt}
	\label{time_compare}%
\end{figure*}

\noindent\textbf{Training Time Comparison:}  
The training times of these models to reach convergence on various frameworks were compared, with results displayed in Figure \ref{time_compare}. 
It indicates that frameworks generally require less training time compared to the origin models. 
This trend is especially prominent in TGL, where it consistently outperforms the original on most datasets and models. 
DistTGL, as a distributed version of TGL, despite having lower evaluation metrics than TGL, shows a reduction in training time. 
SPEED exhibits a significantly longer training time, with up to 10 times longer than TGL, although this disparity may due to the $top_k=10$ setting in SPEED.

\begin{figure*}
	\centering 
	\hspace{-0.5cm}\includegraphics[width=1\textwidth, angle=0]{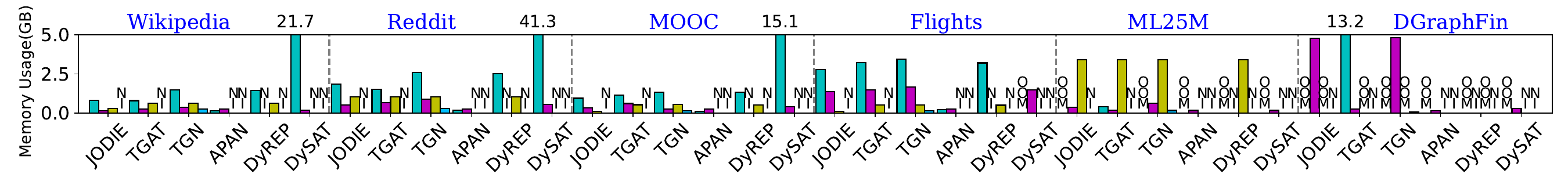}	
    \vspace{-10pt}
	\caption{Comparison of GPU memory usage of models and frameworks, with the GPU memory capacity equals 48GB.} 
	\vspace{-9pt}
	\label{memory_usage_compare}%
\end{figure*}

\noindent\textbf{GPU Memory Usage Comparison:}  
The GPU memory usage during the training of these models on various frameworks was compared, with results displayed in Figure \ref{memory_usage_compare}. 
Generally, the origin models tend to consume more memory compared to frameworks, indicating that frameworks are optimized for dynamic GNN training. 
However, in some cases, SPEED exhibits higher memory usage, possibly due to its graph partitioning approach. 
Frameworks often utilize specific techniques like TGL/DistTGL's T-CSR and SPEED's graph partitioning to accommodate large datasets. 
Despite this, the memory usage of TGL/DistTGL remains lower than that of SPEED, suggesting the efficiency of TGL/DistTGL's T-CSR. 
It's noteworthy that while ML25M and DGraphFin have similar file sizes (as shown in Table \ref{table:datasets}), training DGraphFin requires significantly more memory than ML25M due to GNN sensitivity to node-related factors.
Additionally, these frameworks do not maximize GPU memory usage, with most cases using only up to 4GB.

\subsection{Multi-GPU Scalability} 
\label{scalability} 

\begin{figure}
	\centering 
	\hspace{-0.5cm}\includegraphics[width=0.5\textwidth, angle=0]{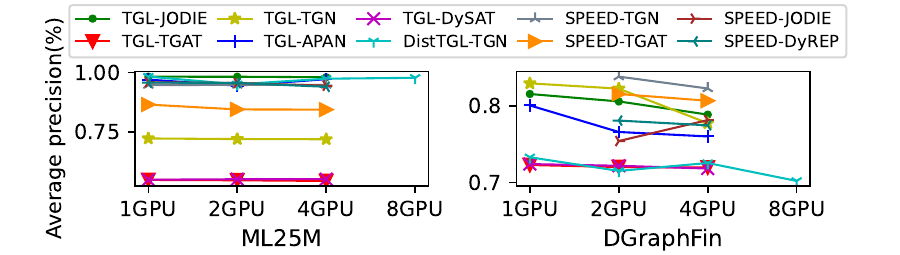}	
    \vspace{-10pt}
	\caption{Average precision for different numbers of GPUs.} 
	\vspace{-10pt}
	\label{Compare_framework_ap}%
\end{figure}

\begin{figure}
	\centering 
	\hspace{-0.5cm}\includegraphics[width=0.5\textwidth, angle=0]{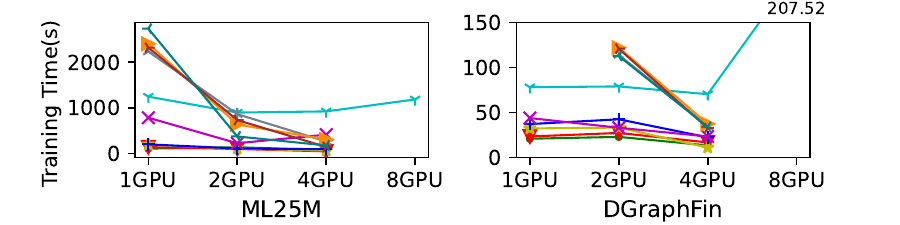}	
    \vspace{-10pt}
	\caption{Average epoch time for different numbers of GPUs.} 
	\vspace{-10pt}
	\label{Compare_framework_time}%
\end{figure}

\begin{figure}
	\centering 
	\hspace{-0.5cm}\includegraphics[width=0.5\textwidth, angle=0]{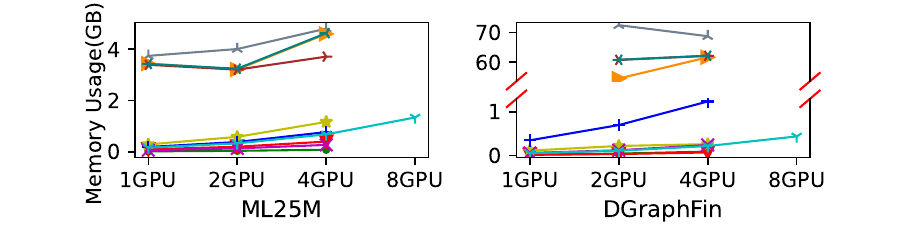}	
    \vspace{-10pt}
	\caption{GPU memory usage for different numbers of GPUs.} 
	\vspace{-17pt}
	\label{Compare_framework_memory}%
\end{figure}

Frameworks generally offer a key advantage over the origin models by incorporating a multi-GPU extension for faster parallel training. 
In this section, we analyze the performance of different frameworks with varying numbers of GPUs. 
We assess performance using 1 GPU, 2 GPUs on one machine, 4 GPUs on one machine, and 8 GPUs on two machines. 
Our analysis is based on the DGraphFin and ML25M datasets, focusing on converged AP, average training time per epoch, and GPU memory usage. 
Results are presented in Figure \ref{Compare_framework_ap}, Figures \ref{Compare_framework_time}, and \ref{Compare_framework_memory}, respectively. 
The lack of data for the 8GPU scenario is due to a lack of multi-machine training support, while the absence for 1GPU is due to out-of-memory issues.

In most scenarios, the AP values demonstrate stability with minimal variation as the number of GPUs increases, suggesting that multi-GPU parallel training has limited impact on model accuracy. 
An exception is seen with TGL, where a potential decrease in AP occurs as the number of GPUs rises, particularly noticeable in the DGraphFin dataset. 
This decline is attributed to TGL's reliance on mini-batch parallelism, which may overlook dependencies within one-time mini-batches, leading to decreased training accuracy when the total size of one-time mini-batches is doubled. 
While DistTGL generally shows resilience to changes in GPU numbers, 
there is a risk of performance deterioration when multiple machines are utilized, 
as evidenced in the DGraphFin results with 8 GPUs spread across two machines. 
This suggests that although DistTGL is relatively unaffected by GPU variations, 
the utilization of multiple machines may introduce accuracy challenges in specific scenarios.

On the other hand, 
the memory usage for models across the three frameworks tends to rise with an increase in the number of GPUs, while the reduction in average per epoch training time is generally not as pronounced, except for specific instances like transitioning from 1 GPU to 2 GPUs in the ML25M dataset. 
This trend may be attributed to the impact of multi-GPU communication, which can contribute to an overall increase in training time.  
Interestingly, DistTGL experiences an escalation in average per epoch training time when moving from 4 GPUs (within one machine) to 8 GPUs (across two machines), notably showing a threefold increase in the case of the DGraphFin dataset. 
This is primarily due to the high cost of communication between multiple machines, resulting in an amplified training time for DistTGL under these configurations.

\subsection{Hyperparameter Settings} 
\label{hyperParameter_setting} 

This section focuses on configuring different batch sizes and GNN layers to assess their impacts on model training performance.  

\begin{figure}
    \centering 
    \hspace{-0.5cm}\includegraphics[width=0.5\textwidth, angle=0]{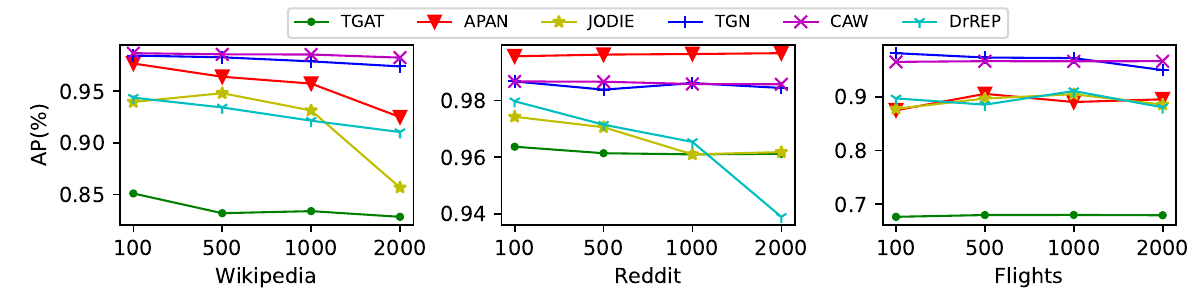}	
    \vspace{-15pt}
    \caption{Average precision for different batch sizes.} 
    \vspace{-10pt}
    \label{batch_size_compare_ap}%
\end{figure}

\begin{figure}
    \centering 
    \hspace{-0.5cm}\includegraphics[width=0.5\textwidth, angle=0]{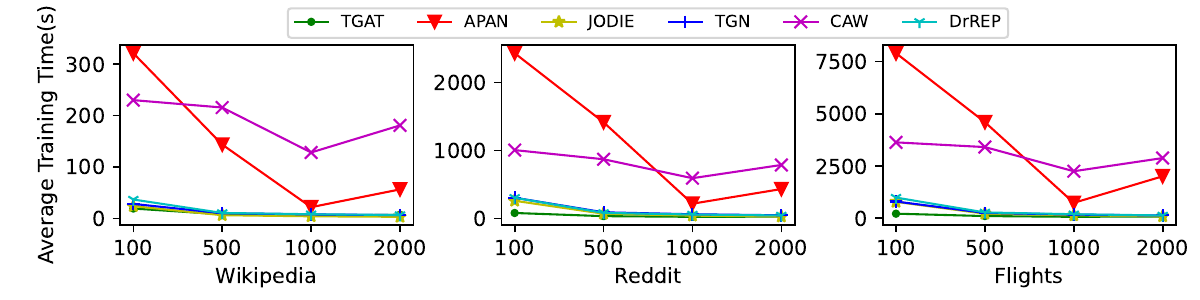}	
    \vspace{-15pt}
    \caption{Training time for different batch sizes.} 
    \vspace{-10pt}
    \label{batch_size_compare_time}%
\end{figure}

\begin{figure}
    \centering 
    \hspace{-0.5cm}\includegraphics[width=0.5\textwidth, angle=0]{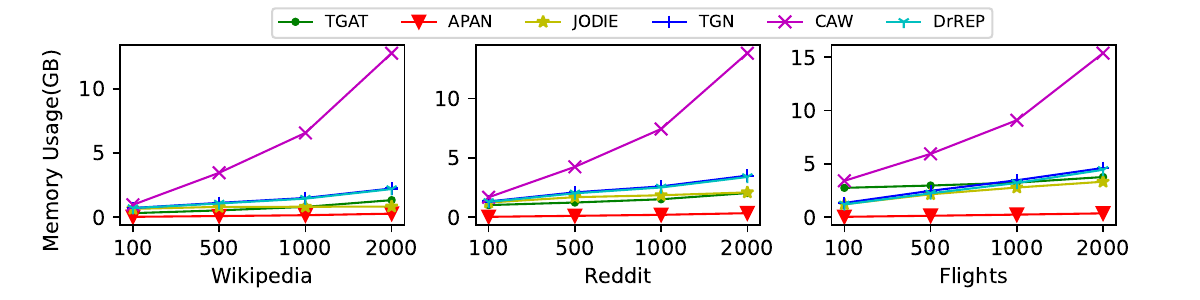}	
    \vspace{-15pt}
    \caption{GPU memory usage for different batch sizes.} 
    \vspace{-17pt}
    \label{batch_size_compare_memory}%
\end{figure}

\noindent\textbf{Impact of Batch Size:} 
We experimented with different batch sizes (100, 500, 1000, and 2000) while maintaining a single GNN layer for the six CTDG models. 
Figure \ref{batch_size_compare_ap} displays the models' AP results for the different batch sizes, revealing that a batch size of 100 consistently yields better performance. 
This suggests that, in general, smaller batch sizes can lead to improved indicator results by enabling more efficient feature learning. 
However, this efficiency may be counterbalanced by longer training times, as depicted in Figure \ref{batch_size_compare_time}.  
Figure \ref{batch_size_compare_time} also indicates that an optimal batch size lies between excessively large and excessively small values, representing a compromise. 
The choice of batch size is typically a trade-off between the number of batches (for smaller sizes) and the computational load per batch (for larger sizes), impacting operational speed. 
Although TGN, DyRep, and JODIE exhibit decreasing trends up to a batch size of 2000, it is anticipated that training time will escalate beyond a certain batch size threshold.  
Additionally, as shown in Figure \ref{batch_size_compare_memory}, there is a direct correlation between batch size and memory usage, with memory requirements increasing as batch size grows.

\begin{figure}[!t]
    \centering 
    \hspace{-0.5cm}
    \includegraphics[width=0.5\textwidth, angle=0]{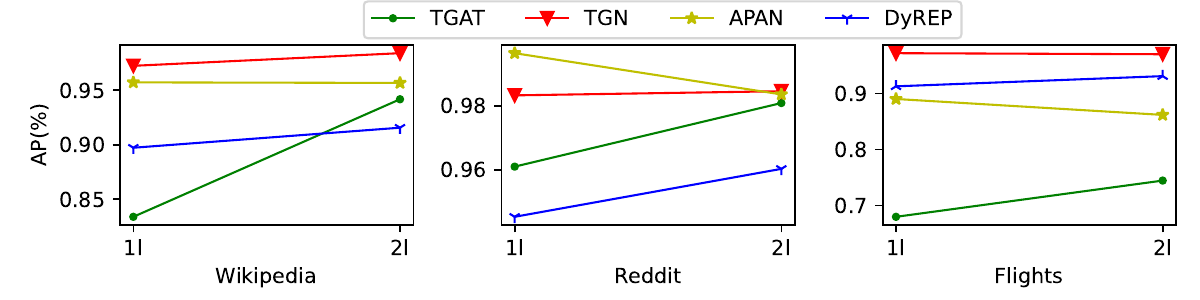}	
    \vspace{-15pt}
    \caption{Average precision for different GNN Layers.} 
    \vspace{-10pt}
    \label{gnn_layer_compare_ap}%
\end{figure}

\begin{figure}
    \centering 
    \hspace{-0.5cm}
    \includegraphics[width=0.5\textwidth, angle=0]{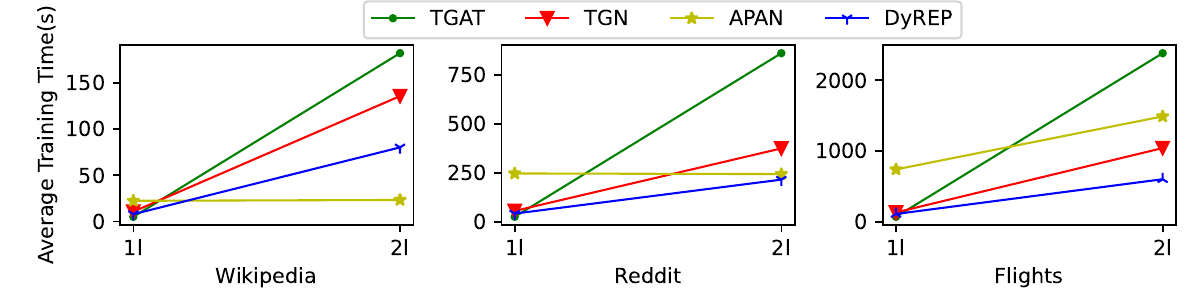}	
    \vspace{-15pt}
    \caption{Training time for different GNN Layers.} 
    \vspace{-10pt}
    \label{gnn_layer_compare_time}%
\end{figure}

\begin{figure}
    \centering 
    \hspace{-0.5cm}
    \includegraphics[width=0.5\textwidth, angle=0]{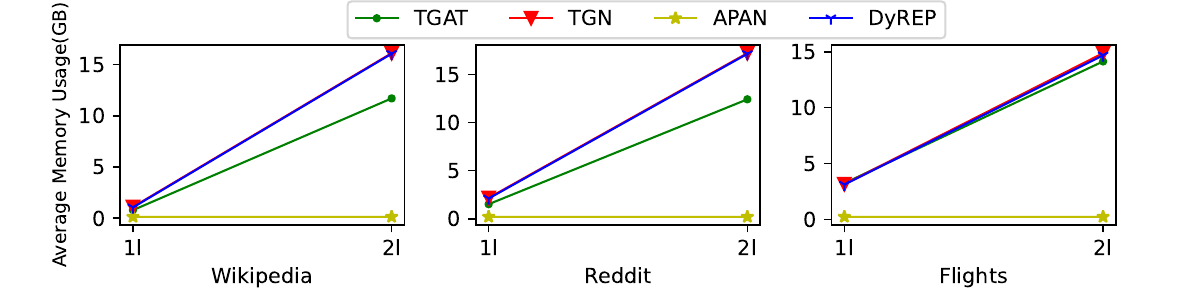}	
    \vspace{-17pt}
    \caption{GPU memory usage for different GNN Layers.} 
    \vspace{-12pt}
    \label{gnn_layer_compare_memory}%
\end{figure}

\noindent\textbf{Impact of GNN Layers:} 
To investigate the impact of different numbers of GNN layers on model performance, we conducted a two-layer experiment with the TGAT, TGN, APAN, and DyREP models. 
JODIE was excluded due to the absence of layer concepts, and CAW was prone to out-of-memory (OOM) issues with two layers. 
The experiments were conducted with a batch size of 1000.  
The evaluation results are presented in Figure \ref{gnn_layer_compare_ap}. 
The findings suggest that increasing the number of GNN layers can enhance evaluation performance. However, as illustrated in Figures \ref{gnn_layer_compare_time} and \ref{gnn_layer_compare_memory}, this improvement comes at the cost of increased training time and memory consumption. 
The objective is to increase the number of GNN layers to improve performance within a reasonable epoch time while mitigating excessive time and OOM concerns.  
Notably, APAN appears less sensitive to the number of layers, as its model architecture is not heavily dependent on this factor. 
Consequently, augmenting the number of layers does not substantially inflate training time or memory usage for APAN.  
While the results for three layers are not presented, TGAT, TGN, and DyREP encountered OOM issues with three layers, whereas APAN did not. 
This discrepancy can be attributed to APAN's model architecture exhibiting minimal correlation with the number of GNN layers.

\subsection{Node Classification Evaluation}
\label{subsec:node_classification}

In this section, we assess the converge accuracy of various dynamic GNN models for the node classification task. 
Specifically, we consider the JODIE, TGN, DyREP, TGAT, and APAN  of CTDG models, as well as the EvolveGCN of the DTDG model. 
These models are evaluated on the Wikipedia, Reddit, and MOOC datasets, using appropriate labels for node classification. 
A single GNN layer with a batch size of 100 is utilized for consistency across all models.  
The results of the AUC and AP metrics are presented in Table \ref{table:node_classification}. 
Interestingly, the performance metrics of these dynamic GNN models exhibit lower values in node classification tasks, particularly evident in the Reddit and MOOC datasets. 
This performance difference may stem from these models being primarily designed for the prevalent link prediction task in dynamic graphs. 
Surprisingly, there is no substantial gap in performance between DTDG and CTDG models in node classification tasks, indicating that both types of dynamic GNN models are similarly effective in this scenario. 
Notably, TGN attains the highest AUC on the Reddit dataset, 
while APAN achieves the highest AUC on the Wikipedia dataset. 
EvolveGCN stands out as the top performer among CTDGs, particularly excelling on the MOOC dataset.

\begin{table}
    \caption{Converge accuracy for node classification task.}
    \vspace{-5pt}
    \renewcommand{\arraystretch}{1.2}
    \scalebox{0.7}{
    \begin{tabular}{| c | c | c | c | c | c | c | c |} 
     \hline
     \multirow{2}{*}{Type} & \multirow{2}{*}{Model} & \multicolumn{2}{c|}{Wikipedia} & \multicolumn{2}{c|}{Reddit}  & \multicolumn{2}{c|}{MOOC}  \\
     \cline{3-8}
      & & AUC & AP & AUC & AP & AUC & AP   \\
     \hline
     \multirow{5}{*}{CTDG} 
     & JODIE & 0.8021 & 0.0125 & 0.6392 & \textbf{0.0033}  & 0.6099 & 0.0218 \\ 
     \cline{2-8}
      &	TGAT &	0.8571 & 0.0124 & 0.6724 & 0.0019 & 0.6311 & 0.0196 \\ 
     \cline{2-8}
      &	TGN-attn & 0.8685 & 0.0137 & \textbf{0.6821} & 0.0020 &  0.5386 & 0.0162 \\ 
     \cline{2-8}
      &	APAN & \textbf{0.8701} & 0.0141 & 0.5590 & 0.0014  & 0.6412 & 0.0302 \\
     \cline{2-8}
      &	DyRep &	0.8357 & 0.0130 & 0.5310 & 0.0010 & 0.6561 & 0.0236 \\ 
     \hline
     \multirow{2}{*}{DTDG} & EvolveGCN-H & 0.6981 & 0.0148 & 0.5921 & 0.0018  & \textbf{0.6792} & 0.0325 \\
     \cline{2-8}
      & EvolveGCN-O & 0.7920 & \textbf{0.0793} &  0.5964 & 0.0021  & 0.6782 & \textbf{0.0372} \\
     \hline
    \end{tabular}
    }
    \vspace{-10pt}
    \label{table:node_classification}
\end{table}

\section{Open Challenges}
\label{sec:challenges}

Despite significant advancements in dynamic GNN models and training frameworks, 
several open challenges continue to exist, placing them in a phase of ongoing exploration 
with limitations in diversity, generality, accuracy, efficiency, and scalability.

\noindent\textbf{Diversity in Application Domains.} 
Dynamic graph representation learning is applied in diverse domains such as social networks, transportation networks, epidemic transmission, and recommendation systems, as discussed in \S\ref{subsec:applications}. 
However, each specific application has unique dynamic graph characteristics, highlighting the need for specialized methods to effectively address the diverse requirements of individual scenarios.

\noindent\textbf{Requirement for a Unified Framework.}  
Existing dynamic graph algorithms utilize various methods to capture time dependencies within graph structures, making it challenging to establish a unified framework. 
While efforts like TGL \cite{TGL} have strived to create a unified graph operator, they typically cover only a restricted range of models. 
Developing a comprehensive unified graph operator capable of encompassing a majority of algorithms is a crucial yet formidable endeavor in the field of dynamic graph learning.

\noindent\textbf{Challenges in Processing Dynamic Graph Updates.} 
While existing dynamic GNN frameworks provide functional support for training modules essential for dynamic graph models, their assistance is constrained, especially in dynamic graph data storage. 
Current frameworks commonly utilize structures like CSR or T-CSR, which encounter difficulties in promptly supporting graph updates to real-time dynamic graph updates. 
This constraint also restricts the adaptability of dynamic graph models.

\noindent\textbf{Challenges in Storing Large-Scale Dynamic Graphs.}  The data volume of dynamic graph data evolving over time is considerably larger than that of static graphs. 
Dynamic GNN models also necessitate storage of additional long-term and short-term memory information related to the evolution of graph data, resulting in significant storage and computational requirements. 
Many existing models are limited in their ability to handle large graphs, and when confronted with such scenarios, they often require substantial GPU resources for processing, indicating scalability challenges. 
As the size of graphs increases, distributed processing becomes a viable solution \cite{DistTGL}. Therefore, there is an urgent need to advance distributed and parallel computing methodologies to effectively manage and process large-scale dynamic graph data.

\noindent\textbf{Inefficient Training Data Extraction.} 
In dynamic GNN training, the extraction of training data involves not only feature data but also historical memory information, which can impede the GNN training efficiency. 
In distributed parallel training frameworks, data extraction also includes data transmission among multiple GPUs and machines, adding to the inefficiency of the process. 
Furthermore, updating this memory information post-training further reduces the effectiveness of data extraction in dynamic GNN training. 
On the other hand, our experiments reveal instances of under-utilization of GPU memory, presenting an opportunity to leverage this available memory space to accelerate the data extraction process.

\noindent\textbf{Challenges in Parallel Training Efficiency.}  
The majority of current frameworks support training on a single machine with single or multiple GPUs but lack distributed environment support across multiple machines. 
This limitation restricts their scalability in parallel training large dynamic graphs. 
Additionally, the increasing complexity of dynamic GNN models and temporal dependencies within dynamic graphs pose further obstacles to parallel training efficiency. 
Current frameworks may 
either neglect dependencies within individual mini-batches to improve parallel training efficiency, 
or concentrate solely on capturing temporal dependencies leading to suboptimal training efficiency.
\section{Conclusion}
\label{sec:conclusion}

This paper provides a comprehensive comparative analysis and experimental evaluation of dynamic GNNs. 
It covers 81 dynamic GNN models with a novel taxonomy, compares 12 dynamic GNN training frameworks, and includes commonly used benchmarks for evaluating dynamic GNNs. 
The paper includes extensive experiments on nine representative models and three frameworks across six standard graph datasets using unified benchmarks and evaluation metrics. 
Performance evaluation covers metrics related to convergence accuracy, training efficiency, and GPU memory usage, considering both single GPU and multiple GPUs scenarios.
The analysis and evaluation results highlight various open challenges in the dynamic GNN field to offer valuable principles for future researchers to enhance the generality, performance, efficiency, and scalability of dynamic GNN models and frameworks.



\bibliographystyle{ACM-Reference-Format}
\balance
\bibliography{cite}

\end{document}